\documentclass[journal]{IEEEtran}
\usepackage[printonlyused]{acronym}
\usepackage[style=ieee,backend=biber]{biblatex}
\bibliography{biblio}

\usepackage{multicol}
\usepackage{amsmath,amssymb,amsfonts}
\usepackage{algorithmic}
\usepackage{algorithm}
\usepackage{array}
\usepackage{textcomp}
\usepackage{stfloats}
\usepackage{url}
\usepackage{verbatim}
\usepackage{graphicx}
\usepackage{xcolor}
\usepackage{subfigure}
\usepackage{hyperref}
\usepackage{bm}
\usepackage{mathtools}
\usepackage{graphicx}
\usepackage{multirow}
\usepackage{booktabs}
\usepackage{placeins}
\usepackage{afterpage}
\usepackage{enumitem}
\usepackage{caption}
\renewcommand{\footnoterule}{\vfill\kern -3pt \hrule width 0.4\columnwidth \kern 2.6pt}
\usepackage{algorithm}
\usepackage{algorithmic}

\newlength\myindent
\newcommand\bindent{%
  \begingroup
  \setlength{\itemindent}{\myindent}
  \addtolength{\algorithmicindent}{\myindent}
}
\newcommand\eindent{\endgroup}

\newcommand{\cd}[1]{{\color{red}#1}} 

\colorlet{red}{black}

\def\BibTeX{{\rm B\kern-.05em{\sc i\kern-.025em b}\kern-.08em
    T\kern-.1667em\lower.7ex\hbox{E}\kern-.125emX}}

\begin{document}

\title{Graph Neural Networks as an Enabler of Terahertz-based Flow-guided Nanoscale Localization over Highly Erroneous Raw Data}

\author{Gerard Calvo Bartra, Filip Lemic\IEEEauthorrefmark{1}, Guillem Pascual, Aina Pérez Rodas, Jakob Struye, \\Carmen Delgado, Xavier Costa P\'erez\vspace{-7mm}
\IEEEcompsocitemizethanks{\IEEEcompsocthanksitem\IEEEauthorrefmark{1}Corresponding author.}
\thanks{G. Calvo, F. Lemic, G. Pascual, C. Delgado, and X. Costa are affiliated with the i2Cat Foundation, Spain, email: \{name.surname@i2cat.net\}. X. Costa is also affiliated with NEC Laboratories Europe, Germany and ICREA, Spain.}
\thanks{G. Calvo, G. Pascual, and A. Perez are affiliated with the Polytechnic University of Catalonia, Spain, email: \{name.surname\}@upc.edu.}
\thanks{J. Struye is affiliated with the University of Antwerp - imec, Belgium, email: \{jakob.struye@uantwerpen.be\}.}
\thanks{Manuscript received July 10, 2023; revised XXX.}}

\markboth{IEEE DRAFT}
{Shell \MakeLowercase{\textit{et al.}}: A Sample Article Using IEEEtran.cls for IEEE Journals}

\maketitle

\begin{abstract}

Contemporary research advances in nanotechnology and material science are rooted in the emergence of nanodevices as a versatile tool that harmonizes sensing, \cd{computing,} wireless communication, data storage, and energy harvesting. 
These devices hold promise in precision medicine, offering novel pathways for disease diagnostics, treatment, and monitoring within the bloodstreams. 
Ensuring precise localization of events of diagnostic interest, which underpins the concept of flow-guided in-body nanoscale localization, would intuitively provide an added diagnostic value to the detected events.
\cd{Raw data generated by the nanodevices is pivotal for this localization and consist of an event detection indicator and the time elapsed since the last passage of a nanodevice through the heart.}
\cd{The communication and energy constraints of the nanodevices lead to intermittent operation and unreliable communication, intrinsically affecting this data.}
This posits a need for comprehensively modelling the features of this data. 
These imperfections also have profound implications for the viability of existing flow-guided localization approaches, which are ill-prepared to address the intricacies of the environment. 
Our first contribution lies in an analytical model of raw data for flow-guided localization, dissecting how communication and energy  capabilities influence the nanodevices' data output. 
This model acts as a vital bridge, reconciling idealized assumptions with practical challenges of flow-guided localization.
Toward addressing these practical challenges, we also present an integration of \acp{GNN} into the flow-guided localization paradigm. 
\acp{GNN}, reinforced by the adaptability and resilience of \acp{HGT}, excel in capturing complex dynamic interactions inherent to the localization of events sensed by the nanodevices. 
Our results highlight the potential of \acp{GNN} not only to enhance localization accuracy but also extend coverage to encompass the entire bloodstream.
\end{abstract}

\vspace{-2mm}


\acrodef{GNN}{Graph Neural Network}
\acrodef{GMM}{Gaussian Mixture Model}
\acrodef{HGT}{Heterogeneous Graph Transformer}
\acrodef{GAT}{Graph Attention Network}
\acrodef{HGT}{Heterogeneous Graph Transformer}
\acrodef{CE}{Cross Entropy}
\acrodef{WaB}[W\&B]{Weights and Biases}
\acrodef{SotA}{State of the Art}
\acrodef{ML}{Machine Learning}
\acrodef{THz}{Terahertz}
\acrodef{ZnO}{Zinc-Oxide}
\acrodef{SINR}{Signal to Interference and Noise Ratio}
\acrodef{NN}{Neural Network}
\acrodef{ML}{Machine Learning}
\acrodef{IMU}{Inertial Measurement Unit}
\acrodef{RF}{Radio Frequency}
\acrodef{ECDF}{Empirical Cumulative Distribution Function}
\acrodef{KL}{Kullback-Leibler}
\acrodef{MW}{Mann-Whitney}
\acrodef{BVS}{BloodVoyagerS}
\acrodef{MSE}{Mean Squared Error}
\acrodef{GPS}{Global Positioning System}
\acrodef{DFS}{Depth-First Search}
\acrodef{GCN}{Graph Convolutional Network}
\section{Introduction}

Research advances in nanotechnology are propelling the development of nanoscale devices with intertwined sensing, computing, and data and energy storage capabilities~\cite{jornet2012joint,senturk2022internet}. 
These miniature devices hold potential for transformative applications in the realm of precision medicine~\cite{abbasi2016nano,emerich2003nanotechnology}. 
Some applications within this realm envisage the nanodevices' deployment within the intricate vascular highways of the human body, where their physical dimensions must closely mirror those of the red blood cells, typically measuring less than 5 microns. 
Nanodevices utilize energy harvesting from physiological sources, such as heartbeats and ultrasound waves, using nanoscale components such as ZnO nanowires~\cite{jornet2012joint,gao2007nanowire}, allowing them to operate in challenging environments. 
Consequently, these nanodevices traverse the bloodstream in a passive, unobtrusive manner.

Contemporary advances in material science, notably targeting graphene and its derivatives~\cite{jornet2013graphene,abadal2015time}, have created possibilities for nanoscale wireless communication at \ac{THz} frequencies (i.e., 0.1-10 THz)~\cite{akyildiz2008nanonetworks,lemic2021survey}. 
The inclusion of wireless communication capabilities enables two-way communication between the nanodevices and the external world~\cite{dressler2015connecting}. 

Nanodevices with integrated communication abilities are facilitating a number of sensing and actuation-based applications in the bloodstream~\cite{muzykantov2011targeted}.
This is because the bloodstream is a common and relevant target for medical diagnostics of ischemia, inflammation, thrombosis, myocardial infarction, bleeding disorders, vascular and pulmonary maladies, stroke, and other conditions primarily involving its key components (i.e., heart, blood, and vessels). 
Moreover, many diseases and conditions localized outside the vascular compartment involve the bloodstream (e.g., diabetes, tumors, pneumonia, asthma, neurological and gastrointestinal diseases). Finally, as stated in~\cite{muzykantov2011targeted}, the bloodstream is a natural route of drug delivery to all targets—tumors, neural systems, and glands. 

Additionally, communication-enabled nanodevices serve as a foundation for flow-guided in-body nanoscale localization within the bloodstreams~\cite{lemic2021survey}. 
Flow-guided localization could enable the association of a location to an event detected by a nanodevice, offering advantages such as non-invasiveness, early and precise diagnostics, and cost reduction~\cite{lemic2022toward,gomez2022nanosensor}.
Its realization is contingent upon addressing a fundamental challenge of the inherently dynamic and complex nature of the human body and bloodstream, rendering the raw data for flow-guided localization highly erroneous~\cite{lopez2023toward,lemic2023insights}. 
The nanodevices require transmitting critical information to the outside world, but communication at the nanoscale poses difficulties, irrespective of the nanocommunication approach employed, as stated in~\cite{stelzner2017function}. 
The erroneous raw data demands a comprehensive approach for modelling and understanding the stochastic nature of this data, regardless of the specific biological event or application at hand.

In response to the challenge, we propose an analytical model that captures the generic features of raw data for flow-guided localization. 
The model accounts for the mobility of nanodevices within the bloodstream and accommodates in-body communication and the energy constraints that characterize contemporary flow-guided localization approaches. 
It offers a unified framework for understanding the challenges that stem from the unreliable nature of data acquisition, independent of the particular biological event or nanodevice type.

The evaluation methods that currently exist for flow-guided localization have mainly focused on specific aspects, like nanodevice mobility, resulting in evaluations that offer only a limited reflection of the complexity and uncertainty that characterizes real-world scenarios~\cite{simonjan2021body,gomez2022nanosensor,lemic2022toward}. 
This observation has been recognized by López~\emph{et al.}~\cite{lopez2023toward}, where the authors provided a simulator that considers multiple factors to enhance the realism of such assessments. These factors encompass the mobility of nanodevices, in-body \ac{THz} communication between nanodevices and the external world, and energy-related and technological constraints at the nanodevice level. 
This approach provides a more comprehensive and realistic understanding of the performance of \ac{SotA} flow-guided localization methods~\cite{lemic2023insights}. 
Specifically, it paints a stark picture of the limitations of these methods in terms of localization accuracy and coverage. 
The unreliability of THz communication between nanodevices and the outside world, compounded by the intermittent nanodevice operation due to energy harvesting, represents a formidable barrier.

In light of this challenge, we propose adopting \acfp{GNN} as an enabler of flow-guided localization over highly erroneous raw data. 
\acp{GNN}, supported by the flexibility and resilient operation of \cd{\acfp{HGT}~\cite{hu2020heterogeneous}}, excel in modeling the intricate and dynamic interactions within heterogeneous graphs~\cite{zhou2020graph}, making them a powerful tool for tackling the challenges posited by the localization of events sensed by nanodevices in the bloodstream.
Our results, derived using the simulator from López~\emph{et al.}~\cite{lopez2023toward}, showcase the superiority of the proposed \ac{GNN} approach over existing \ac{SotA} proposals, such as those found in~\cite{gomez2022nanosensor} and~\cite{lemic2023insights}. 
This superiority is evident in enhanced coverage, allowing for event localization throughout the entire bloodstream and a notable reduction in localization errors. 
However, our findings also bring to the fore the challenges associated with region classification accuracy, both for the \ac{GNN} approach and for baseline methods, stemming from imbalanced and erroneous raw data inherent to the scenario. 
This underscores the pressing need for alternative strategies to enhance accuracy, primarily focusing on introducing additional on-body anchors to support localization and advance the state of flow-guided localization.
For the proposed \ac{GNN} approach, we show that such extensibility to multi-anchor systems is possible, and the resulting accuracies demonstrate the promise of multi-anchor \ac{GNN}-based flow-guided localization.

{\color{red}
The rest of this article is structured as follows. 
Section~\ref{sec:related_work} provides an overview of related efforts.
Specifically, we provide an overview of in-body \ac{RF}-based localization approaches with a particular focus on flow-guided localization considered in this work.
We also discuss the limitations of current performance assessments of flow-guided localization.
Section~\ref{sec:model} presents our analytical model of raw data for flow-guided localization, while Section~\ref{sec:GNN_loc} outlines the proposed approach for GNN-based flow-guided localization.
The evaluation setup and performance assessment of our contributions are given in Sections~\ref{sec:comparison} and~\ref{sec:results}, respectively.
We discuss the limitations of our work and outline promising future research directions in Section~\ref{sec:discussion}, and conclude the article in Section~\ref{sec:conclusion}.
}

{\color{red}
\section{Related Works}
\label{sec:related_work}

\subsection{In-body Flow-guided THz-based Nanoscale Localization}

\begin{figure*}[!t]
	\centering
	\includegraphics[width=0.69\linewidth]{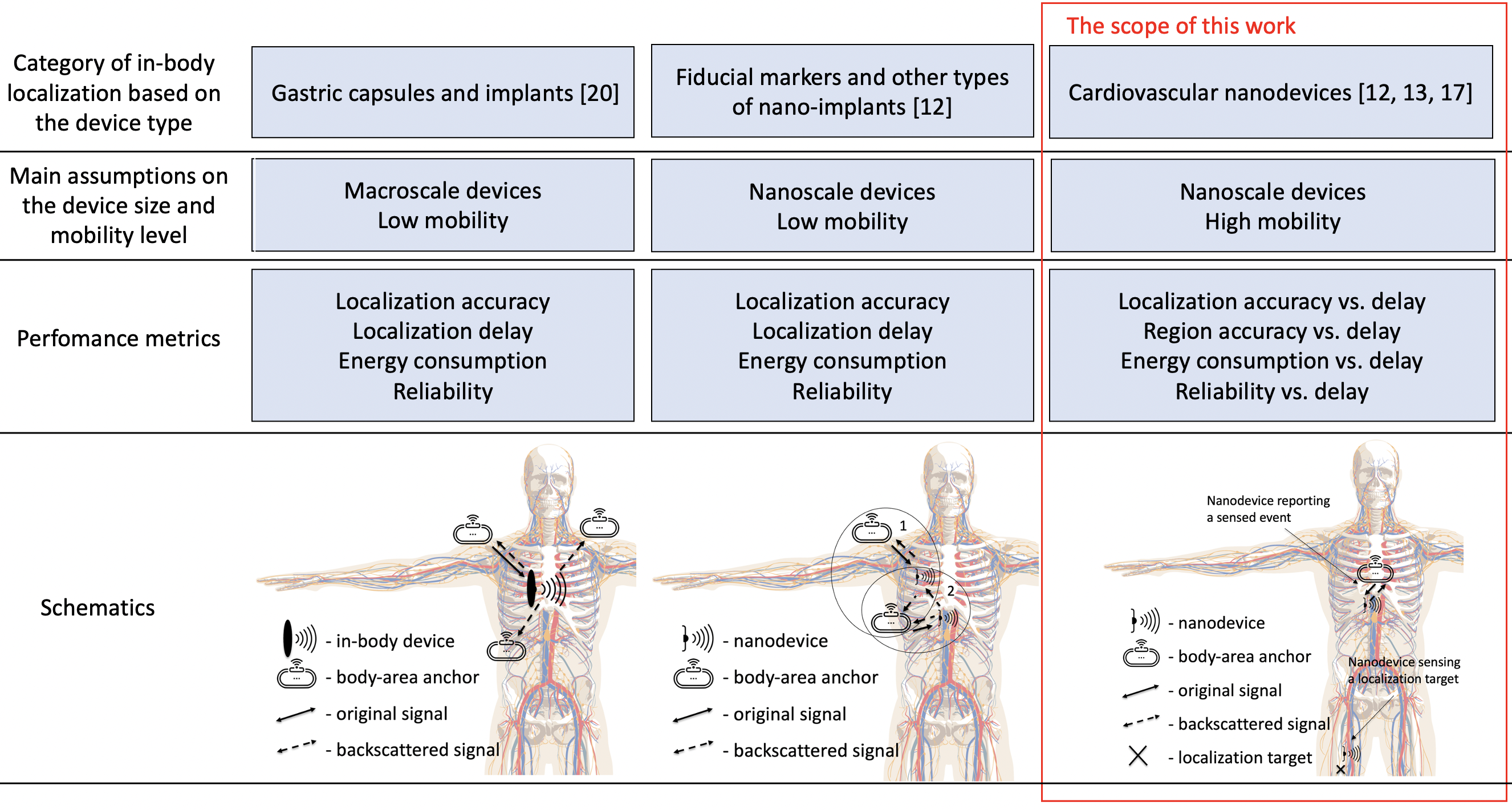}
	\vspace{-1mm}
	\caption{Categorization of RF-based in-body localization approaches, corresponding applications, their requirements, and performance metrics.}
	\label{fig:categorization}
	\vspace{-4mm}
\end{figure*}

\ac{RF}-based localization methods targeting in-body applications can be grouped based on the specific applications they enable, as shown in Figure~\ref{fig:categorization}. 
Intuitively, localization of in-body devices that are movable or nomadic within the body is required; otherwise, their positions may be deduced upon deployment. 
These in-body devices are primarily designed to support three types of applications~\cite{vasisht2018body}.

The first type involves localizing macroscale devices within the body, specifically for tracking gastric capsules and implants. It should be noted that there is a diagnostic benefit in relating measurements of the gastrointestinal system with locations at which they were taken by the specific capsule, as well as a benefit in detecting the movement of implants away from their intended deployment locations.
These devices do not typically feature nanoscale dimensions, allowing the usage of \ac{RF} signals within sub-6GHz frequency bands. 
Either their mobility is low, such as in the case of the gastrointestinal system, where their movement rate may be a few centimeters per hour, or there may be ideally no mobility in the case of implants.
Consequently, localization requirements are less stringent than the other two categories depicted in Figure~\ref{fig:categorization}.
A significant advantage is that there are no strict size constraints for these devices, enabling the utilization of batteries and, hence, not experiencing intermittent behavior. 
A representative example of this approach is the work in~\cite{vasisht2018body}, where the authors localized a static backscattering diode within the body by employing out-of-band aliasing of signals transmitted by an out-of-body anchor operating at a central frequency of 1 GHz.
This approach delivers commendable accuracy, with centimeter-level localization accuracy.

The second category focuses on localizing nanoscale devices with low mobility. 
These devices find applications in tracking fiducial markers (n.b., providing accurate tumor or organ location that moves concerning surrounding anatomy) and various small-scale implants.
While some of these applications can be enabled using macroscale devices~\cite{vasisht2018body}, often unlocking their full potential necessitates the usage of nanoscale entities, especially in scenarios like early targeted treatment of small-scale tumors.
Consequently, we distinctly classified this category in Figure~\ref{fig:categorization}.
A recent effort in this domain is~\cite{lemic2022toward}, where the focus is on densely deployed nanodevices that move through the bloodstream passively. To address this scenario, the authors propose an iterative localization concept. 
This concept involves initially localizing nanodevices closer to the body surface with the support of on-body anchors. 
Subsequently, these localized nanodevices serve as anchors and relays for localizing nanodevices deeper within the body. 
To mitigate the size constraints within the bloodstream, the authors assume energy-harvesting nanodevices operating at THz frequencies. 
This approach could extend to localizing nanoscale implants within the body.  
However, it is also important to acknowledge that addressing the associated challenges, such as stringent latency-related constraints for multi-hop communication, requires further research.

Both of the above categories aim to localize (nano)devices within the body, similar to more traditional indoor localization, where the aim is to find the unknown location of a device (e.g., smartphone) within an indoor environment.  
As a result, evaluation methodologies applicable to traditional indoor localization can also be adapted for in-body (nano)device localization.
For instance, the EVARILOS Benchmarking Methodology~\cite{van2015platform} measures point accuracy (i.e., the Euclidean distance between the true and estimated locations), latency, energy consumption required for localizing the device, and reliability of localization (i.e., probability of reporting a location estimate upon request), as illustrated in Figure~\ref{fig:categorization}.

The third category is the flow-guided localization considered in this work. 
In this context, the objective shifts from localizing the nanodevices to detecting and localizing specific events within the body (cf., Figure~\ref{fig:categorization}). 
While~\cite{lemic2022toward} can conceptually support this type of scenario, the primary representatives of flow-guided localization are~\cite{gomez2022nanosensor} and~\cite{simonjan2021body}.
These approaches leverage \ac{ML} models to distinguish regions through which a nanodevice passed during a single circulation through the bloodstream. 
In~\cite{simonjan2021body}, this is achieved by tracking distances a nanodevice covers during its circulations through the bloodstream, using a conceptual nanoscale \ac{IMU}.
This approach introduces challenges related to \ac{IMU}-generated data storage and processing at the nanodevice and issues tied to the vortex flow of blood, which can adversely affect the accuracy of \ac{IMU} readings. 
In~\cite{gomez2022nanosensor}, these challenges are addressed by tracking the time required for each circulation through the bloodstream. 
The captured distance or time is then reported to a beaconing anchor using short-range THz-based backscattering at the nanodevice level.
In these flow-guided localization approaches, the focus is not on achieving precise point localization of the target event but on detecting the body region, which contrasts~\cite{lemic2022toward} where the authors explore the potential for point localization of a nanodevice at any location through a multi-hop approach.

Enhancing region detection accuracy and reliability can generally be achieved by increasing the number of circulations the nanodevices make through the bloodstream. 
However, this also increases energy consumption.
Hence, in the evaluation of flow-guided localization approaches such as~\cite{lemic2022toward,gomez2022nanosensor,simonjan2021body}, the relevant performance metrics such as the point and region estimation accuracies, reliability, and energy consumption should be considered as a function of the application-specific delay allowed for localizing diagnostic events (cf., Figure~\ref{fig:categorization}).

\subsection{Evaluation of THz-based Flow-guided Localization}

We are drawing insights from indoor localization to evaluate the performance of THz-based flow-guided localization. Specifically, research findings about assessing indoor localization performance can be extended to facilitate an objective and standardized assessment of flow-guided localization, as proposed in~\cite{lopez2023toward}.
One of the pioneering efforts in this domain is the EU EVARILOS project, which aimed to establish a robust performance assessment methodology for RF-based indoor localization~\cite{van2015platform}.
This methodology included various evaluation scenarios designed to capture the performance of assessed solutions across diverse metrics like localization accuracy, latency, and energy consumption. 
The project also provided a web platform with raw data that could be used as input to evaluate indoor localization solutions.
Additionally, the project aimed to assess and mitigate the potential detrimental effects of \ac{RF} interference on the performance of the evaluated solutions.
An added milestone was creating a web platform populated with raw data that could be fed into an indoor localization solution, streamlining the performance assessment process for these solutions.
Similar approaches were adopted in the NIST PerfLoc project~\cite{moayeri2016perfloc}, which extended the evaluation to include not only \ac{RF}-based methods but also \ac{IMU}-based, \ac{GPS}-supported, and other hybrid approaches.
The IPSN/Microsoft Indoor Localization Competition~\cite{lymberopoulos2015realistic} enabled the back-to-back evaluation of different indoor localization approaches under the same conditions.

The lessons derived from these and subsequent endeavors underscore that the performance comparison of various indoor localization approaches can be conducted objectively by adhering to a shared evaluation methodology. 
This methodology encompasses consistent deployment environments, evaluation scenarios, and performance metrics. 
The availability of standardized raw data further simplifies the evaluation process, as these datasets can be utilized as inputs for various indoor localization solutions. 
Importantly, these initiatives shed light on the vulnerabilities of \ac{RF}-based indoor localization, revealing that self-interference and interference from neighboring \ac{RF}-based systems operating within the same frequency band can degrade its performance.

In the current context, the evaluation of flow-guided localization approaches, as evidenced in~\cite{gomez2022nanosensor} and~\cite{simonjan2021body}, has been simplified, primarily considering the mobility of nanodevices. 
Consequently, these evaluations may overlook many effects related to wireless communication, such as RF interference and constraints stemming from energy-harvesting, which result in intermittent nanodevice operation~\cite{jornet2012joint}.
Additionally, it is worth mentioning that~\cite{lemic2022toward} conducted a limited performance evaluation that assessed the number of nanodevices required to localize a nanodevice at any location within the body through a multi-hop approach. 
However, these evaluations lack realism and objective methodology, and their subjective evaluation methodologies limit their accuracy.
This limitation was recognized in~\cite{lopez2023toward}, where assessments were made more realistic by considering nanodevice mobility, in-body nanoscale THz communication between nanodevices and the external environment, energy constraints, and other technological limitations (e.g., pulse-based modulation~\cite{jornet2014femtosecond}) of the nanodevices.
The authors followed by demonstrating relatively poor accuracy of the \ac{SotA} solutions in practical scenarios with the high level of realism yielded by the simulator~\cite{lemic2023insights}. 
This poor accuracy was due to unreliable THz communication between in-body nanodevices and on-body anchors and intermittent operation of the nanodevices due to energy harvesting.
} 

\section{Flow-guided In-Body Nanoscale Localization Overview and Framework}
\label{sec:model}

\begin{figure}[!t]
\centerline{\includegraphics[width=\columnwidth]{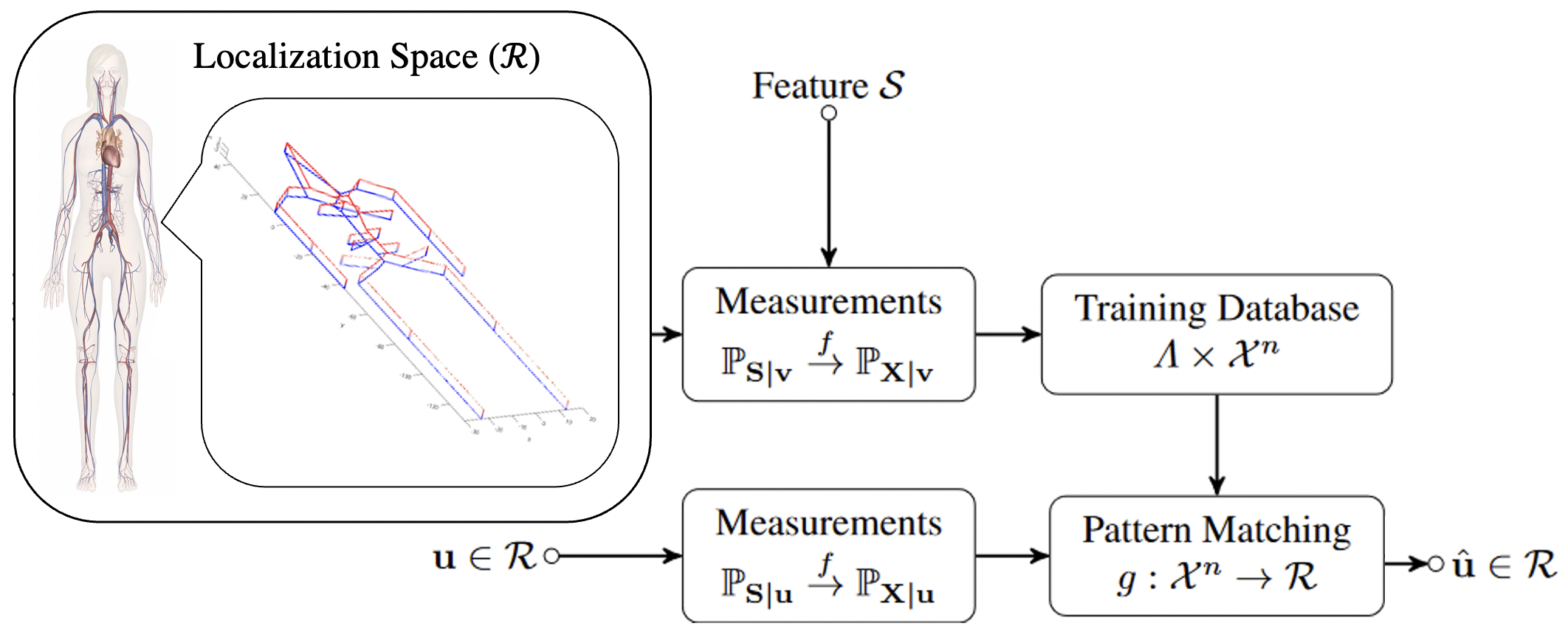}}
\caption{Flow-guided in-body nanoscale localization framework.}
\label{fig:diagrama}
\end{figure}

Our first aim is to analytically capture the peculiarities of raw data for practical flow-guided localization. 
Toward developing the analytical model, we consider a flow-guided localization framework and notations as depicted in Figure~\ref{fig:diagrama}, which was preliminary presented in~\cite{pascual2024math} (n.b., a conference paper).
The framework is adapted from~\cite{behboodi2016mathematical}, where an analytical framework for more traditional fingerprinting-based indoor localization~\cite{caso2019vifi} is proposed. 
Within this approach, a distinct feature of an environment is selected as the basis for creating the fingerprint. 
In the context of flow-guided localization, the environment is an entire bloodstream, modeled as a set of potential cardiovascular paths a nanodevice might pass through in each iteration.

{\color{red}The utilized mathematical notations are summarized in Table~\ref{tab:symbols}.
The chosen feature that is measured and is characteristic of a location is denoted as $S$ and belongs to the feature space $\mathcal{S}$.
In this work and generally, in flow-guided localization, the used measured features are the event detection indicator and the time elapsed since the last passage of a nanodevice through the heart.
Consecutive observations of the measured feature, represented as \(S = (S_1, . . . , S_m) \in \mathcal{S}\), form a random vector that is linked to the location \(u\) through the conditional probability \(P_{S\mid u}\). 
Raw data for flow-guided localization is constructed based on the observations of a certain unique feature of each location through a raw data-creating function.}

The subsequent stage involves the creation of a training database through feature measurements \(S\) at various training locations. 
This database serves as a reference for the subsequent location estimation. 
To determine the location of a nanodevice at \(u\), a pattern-matching function \(g\) is utilized. 
By comparing the acquired raw data with the instances stored in the training database, the pattern matching function \(g\) estimates the location based on the closest matching instances. 
The summarized flow-guided localization framework is:

\begin{itemize}[leftmargin=*]
     \item \textbf{Localization space $\mathcal{R}$:} the cardiovascular system, possible detection regions \(\{R_1,....,R_r\}\), each with the iteration times \(\{T_1,...,T_r\}\).
     \item \textbf{Feature $\mathcal{S}$ = Raw data} ($\mathcal{X}$): Circulation time through a cardiovascular path and event bit \((t,b)\).
     \item \textbf{Pattern matching function \(\bm{g}\):} \ac{ML}-based flow-guided localization approach (e.g., the GNN model from Section~\ref{sec:GNN_loc}).
\end{itemize}

\begin{table}[!t]
{\color{red}
    \centering
    \scriptsize
    \caption{Overview of utilized notations.}
    \label{tab:symbols}
        \begin{tabular}{c p{2.55cm} c p{3.1cm}}
        \hline
        \textbf{Symbol} & \textbf{Description} & \textbf{Symbol} & \textbf{Description} \\
        \hline
        $\mathcal{R}$ & Localization space  & $R_i$ & Region \\
        $\mathcal{S}$ & Feature space & $S$ & Feature \\
        $\mathcal{X}$ & Raw data space &  $X$ & Raw data \\
        $\Lambda$ & Training space & $w,u$ & Training/measurement location \\
        $g$ & Pattern matching function & $T_i$ & Iteration times \\    
        $P_{trans}$ & Transmission probability & $P_{det}$ & Detection probability \\
        \hline
        \end{tabular}}
        \vspace{-3mm}
\end{table}

Based on the outlined framework, we aim to analytically derive the conditional probabilities \(P_{S\mid u}\) as a function of:
\begin{itemize}[leftmargin=*]
    \item \textbf{Probability of detection \bm{$P_{det}$}} corresponds to the probability of an event being detected. This parameter encapsulates the intermittent nature of a nanodevice due to energy-harvesting, which might result in a nanodevice not detecting an event of interest despite it passing through the path containing the event if the nanodevice is off. This results in an erroneous event bit $b$.
    \item \textbf{Probability of transmission \bm{$P_{trans}$}} corresponds to the probability of transmitting data to the on-body anchor close to the heart. 
    It incorporates factors such as having sufficient energy for communication, being within the anchor's range, and self-interference between nanodevices. If the data is not properly communicated to the anchor, the iteration time will not reset, resulting in compound iterations.
    
\end{itemize}


\section{Analytical Modelling of Raw Data for Flow-Guided In-body Nanoscale Localization}
\label{sec:prob_distrib}

Based on the notations established previously, raw data for flow-guided localization corresponds to a tuple \(X=(t,b)\) with \(t \in (0,M]\) and \(b \in \{0,1\}\). \(M\) represents the total duration that the nanodevices spend in the bloodstream.
The time elapsed between transmissions is a composition of the travel times across different regions, each accompanied by zero-mean distributed Gaussian variability \(Q\).
This variability accounts for deviations in iteration times due to factors such as turbulent blood flow, short blood pressure variability, and similar biological factors.
Hence, we model the iteration times as a combination of deterministic travel times and random perturbations captured as the Gaussian noise.
The raw data for flow-guided localization is then:
\begin{equation}
X=\{(n_1T_1+...+n_rT_r+Q,b)\mid n_i \in \mathbb{N}, b \in \{0,1\}\}.
\end{equation}
An important observation is that a subset of raw data exists that can never occur depending on the region where the event is located.  
This subset corresponds to cases where the event has been detected ($b=1$), but the nanodevice has not passed through the event region.  
Let \(X_i\) represent the set of possible raw data instances for an event in region \(R_i\); then the mentioned subset must be subtracted from the total raw data set \(X\):

\vspace{-3mm}
\begin{equation}
X_i= X \backslash \{(n_1T_1+...+n_rT_r+Q,1)\mid n_i=0\}.
\end{equation}

In other words, $X_i$ is the subset of $X$ that does not contain raw data instances with event bit $b=1$ that have not passed through the cardiovascular path containing the event.

We follow by modeling the effects of imperfect communication and event detection on the raw data for flow-guided localization, resulting in the following: 

\begin{itemize}[leftmargin=*]
    \item \textbf{Compound iterations:} As previously mentioned, if the nanodevice fails to communicate with the anchor, the iteration time will not reset, resulting in a longer duration.
    \item  \textbf{False negatives:} When the nanodevice fails to detect the target, although it has passed through the affected region.
\end{itemize}

Given that we are dealing with a binary event bit $b$, we distinguish cases where the event is detected and vice-versa, and calculate the probabilities of interest as follows.

\subsection{Case 1: Event Detected (b=1)}
Suppose that the event to be detected is in region \(R_j\), leading to the following expressions:

\vspace{-3mm}
\begin{equation}
P(X=( n_1T_1+...+n_rT_r,1)\mid R_j)=P(\chi_{(n,1)}\mid R_j),
\end{equation}

where \(\chi_{(n,1)}\) refers to this particular raw data instance. 
There are multiple ways in which this raw data instance can be obtained, taking into account that:
\begin{itemize}[leftmargin=*]
    \item There are multiple ways the nanodevice can travel this number of times through each cardiovascular path. This number is given by the number of permutations of the multiset \(\{P_{R_1}^{n_1},...,P_{R_r}^{n_r}\}\), where \(P_{R_i}\) is the probability of the nanodevice traveling through each region.
    The number of permutations corresponds to the expression:
    \vspace{-1mm}
    \begin{equation}
    \binom{n_1+...+n_r}{n_1,...,n_r}= \frac{(n_1+...+n_r)!}{n_1!...n_r!}.
    \end{equation}
    From that it follows:
    \vspace{-2mm}
    \begin{equation}
    P(\chi_{(n,1)}\mid R_j) \propto \binom{n_1+...+n_r}{n_1,...,n_r}P_{R_1}^{n_1}...P_{R_r}^{n_r}.
    \end{equation}

    \item The detection can occur in any iteration through \(R_j\), and once it is detected, the event bit will not change. 
    Let \(P_{d_1}\) be the probability of detecting the event in iteration \(i\), then:
    \begin{equation}
    P_{d_i}=(1-P_{det})^{i-1}P_{det}.
    \end{equation}
\end{itemize}
It is important to consider that the communication was only successful during the last iteration and not in the previous ones. 
Otherwise, the time would have reset when the communication was successful. 
Thus, a multiplicative factor is applied to account for this condition, which is defined as follows.
\vspace{-1mm}
\begin{equation}
P_t=(1-P_{trans})^{(n_1+...+n_r-1)}P_{trans}.
\end{equation}
Finally, the total probability of having a certain raw data instance with event bit $b=1$ is:
\vspace{-1mm}
\begin{equation}
P(\chi_{(n,1)}\mid R_j)=\binom{n_1+...+n_r}{n_1,...,n_r}P_{R_1}^{n_1}...P_{R_r}^{n_r}P_t\sum_{i=1}^{n_j} P_{d_i}.
\end{equation}

\subsection{Case 2: Event not Detected (b=0)}

The probabilities for cases where no detection has occurred are the following, assuming an event located in region \(R_j\):
\vspace{-1mm}
\begin{equation}
P(X=( n_1T_1+...+n_rT_r,0)\mid R_j)=P(\chi_{(n,0)}\mid R_j).
\end{equation}

This expression is comparable to the former case, but now the event is not detected in any of the iterations through \(R_j\), implying that the following term is always multiplying.
This term is defined as \(P_{nd}\), corresponding to the probability of not detecting an event in any iteration in which the event could have been detected:

\vspace{-2mm}
\begin{equation}
P_{nd}=(1-P_{det})^{n_j},
\end{equation}
leading to the expression:
\vspace{-2mm}
\begin{equation}
P(\chi_{(n,0)}\mid R_j)=\binom{n_1+...+n_r}{n_1,...,n_r}P_{R_1}^{n_1}...P_{R_r}^{n_r}P_{nd}P_t.
\end{equation}

\subsection{An Example}

\begin{figure}[t]
\centerline{\includegraphics[width=0.82\columnwidth]{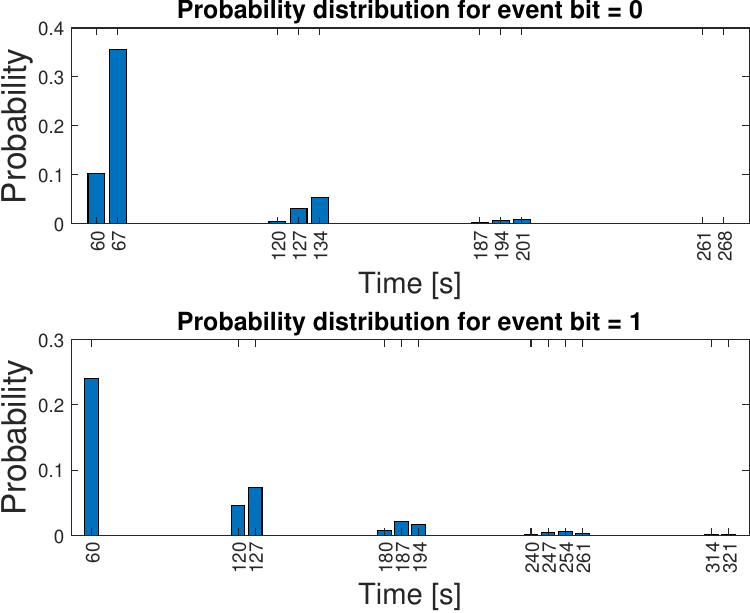}}
\vspace{-1mm}
\caption{Probability distribution for two arbitrary regions with different traveling times (60,67), with probabilities: \(P_{det}=0.7, P_{trans}=0.7\).}
\label{fig:Probd}
\vspace{-2mm}
\end{figure}

As an example, two arbitrary detection regions \((R_1,R_2)\) are considered, with  corresponding traveling times \((T_1,T_2)=(60,67)\)~[sec] and probabilities \((P_{R_1},P_{R_2})=(0.49,0.51)\). 
The probability distributions can be computed using the probabilities (\(P_{R_i}\)) and traveling times (\(T_i\)) of each region, the detection (\(P_{det}\)) and transmission (\(P_{trans}\)) probabilities, the region containing the target event, and the duration of the administration of the nanodevices in the bloodstream. 
From that, it uses a recursive algorithm to account for all possible combinations of iterations with different times, and from there, it computes the probability for every case.
Figure~\ref{fig:Probd} shows the probability distribution assuming a target event in \(R_1\), with $P_{det}=0.7$ and $P_{trans}=0.7$. 
As visible, the highest probabilities correspond to the traveling times of each region. However, there are compound iterations with specific probabilities, such as those corresponding to 120, 127, and 134~s.  
The probabilities become practically unnoticeable after three compound iterations, suggesting that the system converges after several iterations through the bloodstream.

\begin{figure}[t]
\centerline{\includegraphics[width=6.1cm]{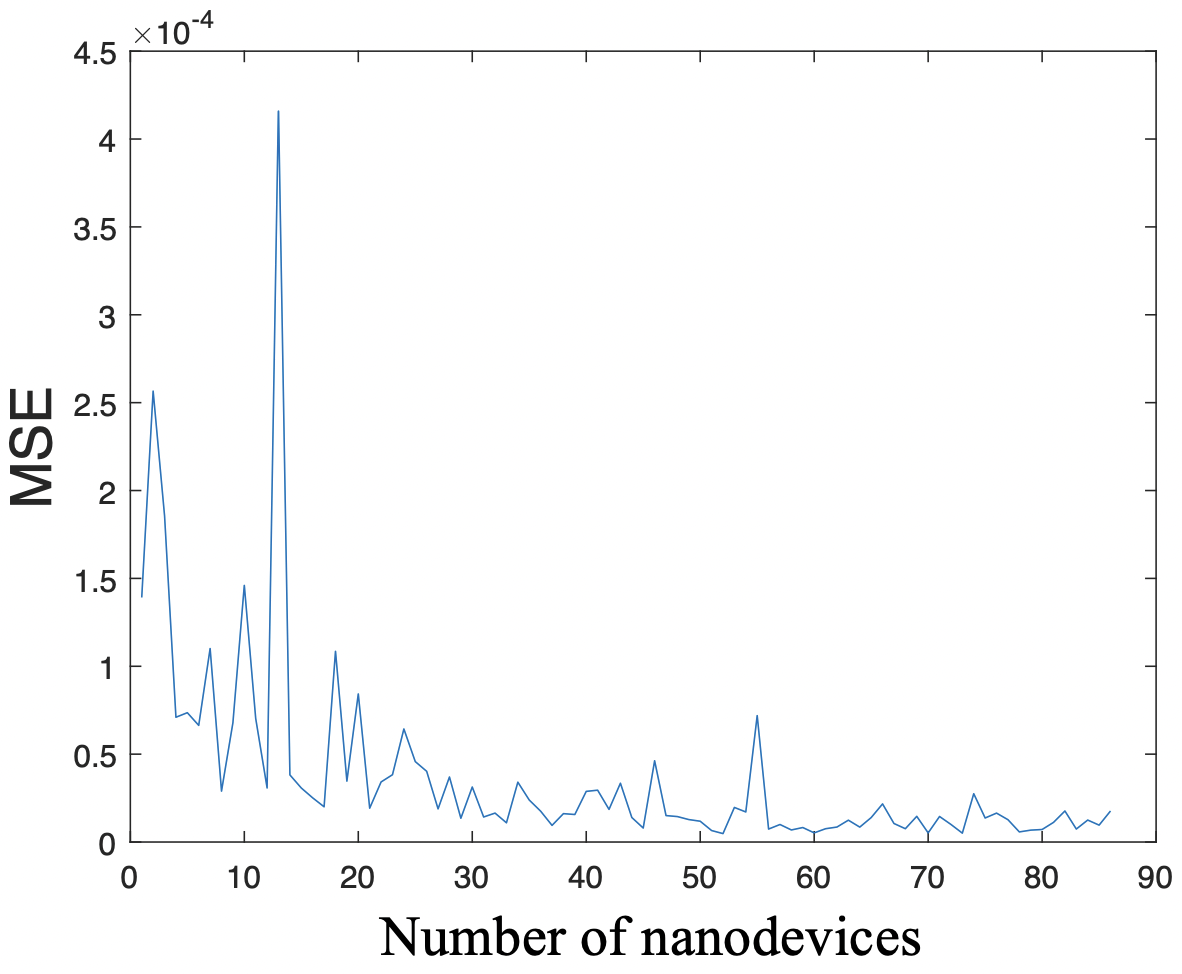}}
\caption{MSE of generated frequencies with respect to the probability distribution as a function of the number of nanodevices.}
\label{fig:mse}
\vspace{-5mm}
\end{figure}

Until now, we have considered a single nanodevice deployed in the bloodstream, although numerous applications envision the administration of many such devices. 
The presence of more nanodevices leads to an augmented volume of data, which can be assessed by plotting accuracy measures against the sample size.
We expect the frequencies to converge to the values of the probability distribution ultimately.  
To discern the convergence rate, we can compute the \ac{MSE} between generated data frequencies and the probability distribution while varying the number of nanodevices.
Example results are depicted in Figure~\ref{fig:mse}, revealing that the \ac{MSE} decreases rapidly with the number of nanodevices, indicating that the distribution of raw data does not change significantly with an increase in the number of nanodevices, apart from increasing the frequency of data transmissions, indicating the utility of the proposed model for modeling the raw data obtained by simultaneously utilizing more than one nanodevice.

\section{GNN-enabled THz-based Flow-guided Nanoscale Localization}
\label{sec:GNN_loc}

\subsection{GNN-enabled THz-based Flow-guided Localization}
\label{subsec:GNN_loc_A}

One secondary objective is to develop a \ac{GNN} model capable of propagating the information from the anchors through the body regions to estimate the event's location. 
{\color{red}
In the context of the flow-guided localization framework presented in Figure~\ref{fig:diagrama}, the GNN model corresponds to the pattern-matching function for correlating the unknown location of a set of observed signal features based on the corresponding raw data set used for training. 
In developing the GNN model, it is essential to note that the bloodstream is a structured and highly connected environment. Hence, it can be represented as a set of edges associated with node coordinates, where each node corresponds to a region of an arbitrary granularity.
In the derivation of our GNN model, we assume the \ac{BVS}-modeled bloodstream with 94-regional granularity~\cite{geyer2018bloodvoyagers}. 
Each edge has a specific length and flow velocity, defining the constraints for the movement of the nanodevices. 
} 

\begin{figure}
\centering
\includegraphics[width=0.97\linewidth]{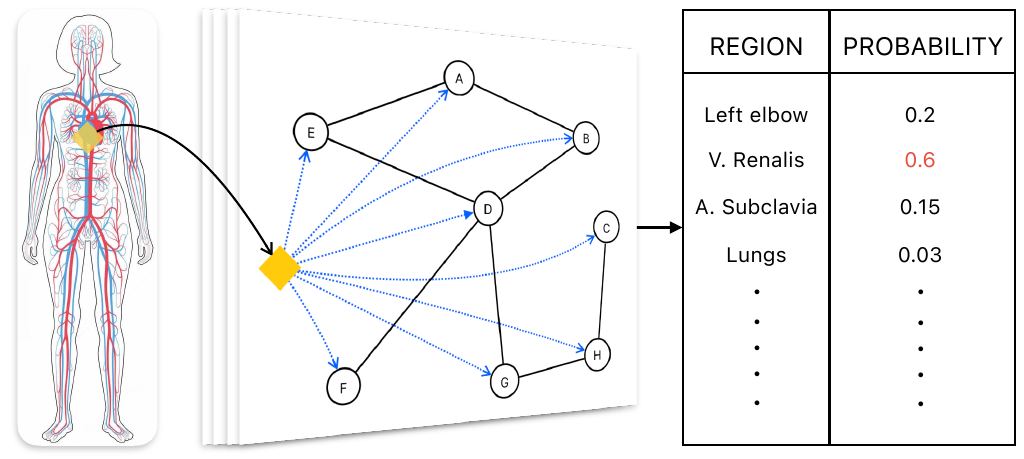}
\vspace{-1mm}
\caption{Overview of GNN-based flow-guided nanoscale localization.}
\label{fig:overview}
\vspace{-5mm}
\end{figure}

\acp{GNN} allow for exploiting relational information in the graph, which is envisioned to facilitate accurate localization of the events in the bloodstream.
The \ac{GNN} proposed in this work leverages region nodes and anchors. 
\cd{As visible in Figure \ref{fig:overview}}, the region nodes hold the information of the regions, such as region type (organ/limb/head, vein, or artery), length, and blood speed. 
The anchors carry the information on the circulation times of the positive bits received from the nanodevices for the localization process. 
Because this data can be of variant length (the anchor receives an undefined number of positive event bits), we create a parameterizable distribution of the bivariate data (i.e., pairs of loop elapsed time and event bit) for each anchor. 
Based on the previously derived analytical raw data insights, we model the circulation time for positive event bits as a \ac{GMM}~\cite{jiang2019gaussian}, where we expect to have a Gaussian cluster for each time a nanodevice fails to communicate to the anchor and runs a second loop in the bloodstream. 
With this approach, the distribution parameters derived using the \ac{GMM} serve as features for the anchors alongside the average number of positive bits received, thus providing a fixed-length feature set for each anchor.

\subsection{GNN Architecture}

\begin{figure}
\centering
\includegraphics[width=\linewidth]{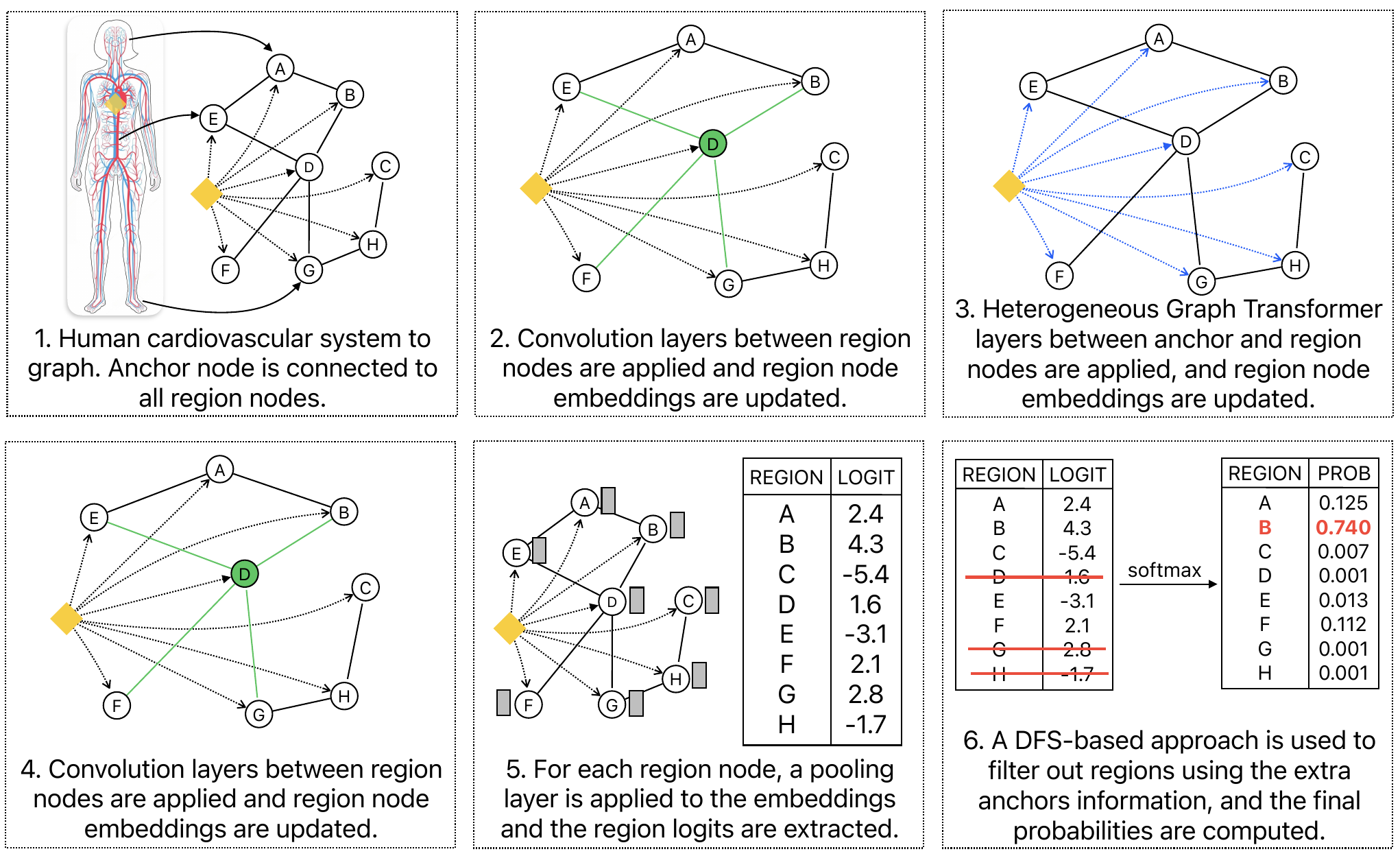}
\vspace{-3mm}
\caption{GNN-based flow-guided nanoscale localization design.}
\label{fig:vision}
\vspace{-5mm}
\end{figure}

The \ac{GNN} architecture employs a comprehensive design paradigm, aiming to leverage the inherent structure of the graph representing the bloodstream. 
\acp{HGT} form the basis of our architecture as they provide versatility to handle the system's multiple types of nodes, i.e., region nodes and anchors.
The \ac{GNN} starts its operations by generating unique embeddings for each node type, as shown in Figure~\ref{fig:vision}. 
It applies a linear transformation and a non-linear activation function (i.e., ReLU) to the initial node features, effectively transforming the information into a latent space of higher dimensionality. 
This initial transformation significantly aids in the propagation and processing of information in the later stages of the model.
The main body of the architecture consists of three principal components: an initial set of convolution layers, a suite of \ac{HGT} layers, and a concluding collection of convolution layers, where each component contributes to the model's efficacy and adaptability.

%

{\color{red}
The primary purpose of the initial convolution layers is to introduce non-linearity and adaptability into the model. 
These layers aggregate information from each region node’s neighbors based on their attention score (relative importance), a measure learned during training. 
The information aggregated is a projection of the node embeddings, information-rich vectors of high dimensions.
The layer employs an attention mechanism that calculates the attention scores, represented by $\alpha_i$ for each neighbor $i$. 
The scores are computed through a specialized process involving transforming the main node's and its neighbor's embeddings.
The embeddings of each node are linearly transformed by a shared weight matrix, and concatenated. 
This concatenated vector is multiplied by a learnable parameter and passed through the LeakyReLU activation to introduce non-linearity, resulting in preliminary attention scores. 
These scores are subsequently normalized across all neighbors of a node using the softmax function to ensure that the sum of $\alpha_i$ values equals 1. 
The normalization makes the main node's embedding a weighted average of its neighbors' embeddings, with weights reflecting their relative importance.
The choice between using the convolution layer \acp{GAT}~\cite{velickovic2017graph} or \ac{GCN} is a hyperparameter. 
This is because, for specific hyperparameter combinations, we have noticed that GCN can outperform GAT layers. 
The GCN is a simpler layer that performs a convolution on the neighbors' embeddings without incorporating the $\alpha$ vector.
}

Following the initial layers, the \ac{HGT} layers are crucial for dealing with the complex interactions between different types of nodes. 
These layers incorporate the information from the anchors to region nodes, enabling the model to capture the dynamic propagation of nanodevices through different body regions.
By dynamically adjusting the importance of different nodes based on the information they are carrying, the \ac{HGT} layers provide a nuanced representation that encapsulates the spatial and temporal aspects of the nanodevices' propagation. 
Our \cd{model} links the anchors to all region nodes, enabling efficient communication between them and eliminating the need for multiple stacked message-passing layers to ensure that information from the anchors reaches all the region nodes.

After the \ac{HGT} layers have processed the information, a set of final convolution layers \cd{are} applied to refine the region nodes' representations. 
Similar to the initial convolution layers, these layers utilize \acp{GAT} to amplify the refined information obtained from the \ac{HGT} layers.
The architecture concludes with a final linear layer applied to the refined representations of the region nodes.
\cd{The output then undergoes a sigmoid activation function to produce the final predictions}, indicating the likelihood of an event occurring in each region.

{\color{red}

\subsection{DFS-based Extension to Multi-anchor Systems}

The \ac{HGT} model outlined above provides initial probabilities for the occurrence of the detected event across all regions in the cardiovascular system.
The data utilized by the \ac{HGT} model is obtained solely by the anchor in the proximity of the heart.
Intuitively, such a system cannot distinguish between left and right counterparts on the body (e.g., limbs). 
In the following, we propose an approach for the utilization of additional strategically positioned on-body anchors for discriminating such regions, intending to refine the accuracy of flow-guided localization further.  
We base our approach on strategically positioning additional anchors so that they can discriminate if an event is detected within cardiovascular paths leading to that anchor from the heart.
Each of these strategically positioned anchors is assumed to be continuously beaconing (n.b., the same assumption as for the anchor positioned in the proximity of the heart) and receiving backscattering responses indicating if the event was detected from the nanodevices.

Intuitively, our extension to multi-anchor systems involves each additional anchor being indicative of events being detected in a sub-region of the body. 
In particular, if all additional anchors yield negative indicators, corresponding regions along their paths are excluded from consideration. 
Conversely, if positive indicators are yielded, consideration is focused solely on the overlapping regions of their paths. 
If additional anchors yield both positive and negative indicators, then the intersection of paths with positive indicators is considered. 
In the absence of any anchors, all regions are considered. 
Additionally, the initial predictions from the HGT model are adjusted based on the refined search area, with regions outside this area being assigned lower probabilities, emphasizing attention on the most probable genomic regions.

In our proposed model, we denote the cardiovascular graph as $G$. 
For each additional anchor $v_i$ in $G$, we calculate the set of regions $A_i$ that nanodevices could traverse to reach $v_i$ from the anchor $s$ in the proximity of the heart.
This is achieved using the \ac{DFS} algorithm:
\begin{equation}
    A_i = \text{DFS}(G, s, v_i).
\end{equation}
The model outputs a binary prediction for each anchor $v_i$, with '1' indicating a likely presence of the event and '0' vice-versa. 
Assume that $S_1$ and $S_0$ are the sets of indices with '1' and '0' predictions. 
The refined set of possible regions $R$ is then determined as follows:

\begin{itemize}[leftmargin=*]
    \item If $S_1$ is empty: 
    \begin{equation}
    R = \text{All regions} \setminus \bigcup_{j \in S_0} A_j
    \end{equation}

    \item If $S_0$ is empty: 
    \begin{equation}
    R = \bigcap_{i \in S_1} A_i
    \end{equation}

    \item If $S_1$ and $S_0$ have elements, it follows: 
    \begin{equation}
    R = (\bigcap_{i \in S_1} A_i) \setminus (\bigcup_{j \in S_0} A_j)
    \end{equation}

    \item If $R$ is empty, consider all regions:
    \begin{equation}
    R = \begin{cases}
        \text{All regions} & \text{if } R = \emptyset \\
        R & \text{otherwise}
    \end{cases}
    \end{equation}
\end{itemize}

In the final step, the logits for regions not in $R$ in the HGT model output are set to a minimal value, effectively ruling them out. 
This process enhances the relative probabilities in the remaining regions, thus focusing the model's prediction only on these areas.
The pseudocode of the HGT with DFS-based multi-anchor extension is given in Algorithm~\ref{alg:pseudocode}.
}

\newcommand{\INDSTATE}[1][1]{\STATE\hspace{#1\algorithmicindent}}
\setlength{\textfloatsep}{4pt}

\begin{algorithm}[!t]
    {\color{red}
    \caption{HGT with DFS-based multi-anchor extension.}
    \label{alg:pseudocode}
    \footnotesize
    \begin{algorithmic}
    \STATE \textbf{INPUT:}
        \bindent 
        \STATE Graph with region and anchor nodes and edges. 
        \STATE Region and anchor nodes have feature vectors.
        \STATE Additional anchor features if available.
        \eindent
        \STATE
    \STATE  \textbf{OUTPUT:}
        \bindent 
        \STATE Graph-level logits with filtering based on additional anchors.  
        \eindent
        \STATE
    \STATE  \textbf{PROCEDURE} HGTWithFiltering:
        \bindent 
        \STATE Initialize base HGT model and Additional Anchor model.
        \bindent
        \STATE
    \STATE \textbf{FUNCTION} DFSAllPaths(start node, target node):
            \INDSTATE[0.3] Recursively explore cardiovascular paths from start to target.
            \INDSTATE[0.3] Return all nodes encountered on these paths.
        \STATE \textbf{END FUNCTION} 
        \STATE
    \STATE \textbf{FUNCTION} CalculateR(additional\_anchors\_out, cover\_sets):
                \INDSTATE[0.3] Calculate the set of allowed region nodes.
                \INDSTATE[0.3] S1: Set of anchors predicting event presence.
                \INDSTATE[0.3] S0: Set of anchors predicting event absence.
                \INDSTATE[0.3] R: Resulting region nodes after filtering using S1 and S0.
                \INDSTATE[0.3] Return R.
        \STATE \textbf{END FUNCTION} 
        \STATE
    \STATE \textbf{FUNCTION} Forward(Graph):
            \INDSTATE[0.3] Compute HGT\_out from passing Graph to HGT model.
            \INDSTATE[0.3] \textbf{IF} additional anchor features are available:
                \INDSTATE[0.45] Compute predictions using the Additional Anchor model.
                \INDSTATE[0.45] Calculate cover sets A for each extra anchor.
                \INDSTATE[0.45] Calculate R using the filtering logic.
                \INDSTATE[0.45] Filter the logits based on R, penalizing regions not in R.
                \INDSTATE[0.3]  \textbf{END IF} 
           \INDSTATE[0.3]  Return the filtered or original HGT\_out as appropriate.
        \STATE \textbf{END FUNCTION} 
        \eindent  
    \eindent 
    \STATE 
    \STATE \textbf{END PROCEDURE} 
    \end{algorithmic}}
\end{algorithm}

\subsection{Model Hyperparameterization and Training}
We trained the \ac{GNN} model using an extensive hyperparameter tuning process and a robust training procedure in PyTorch Geometric to optimize the flow-guided localization of nanodevices within the bloodstream.
The model was trained on an exhaustive simulated training dataset in the format described in Section~\ref{subsec:GNN_loc_A}.
The dataset comprised 1200 graphs, each representing a simulation. Along with node and anchor features, the graphs included a label for the true event region. 
Only simulations with more than one positive bit were selected, which reduced the data for some labels.

Given the data imbalance, the training aimed at minimizing the loss and optimizing the F1 score, which measures the balance between precision and recall and is especially useful for such imbalanced datasets.
By optimizing for the F1 score, the model is encouraged to correctly predict the positive class, i.e., correctly identify the regions where the event is located, which is the model's primary objective.
We standardized the region features of the dataset for enhanced training robustness. 
The model employed the \ac{CE} loss function with sum reduction to aggregate the loss, ensuring the learning process considers all data points. 
The loss was weighed inversely proportional to the class frequencies to account for the class imbalance. 
We used the Adam optimizer, while the hyperparameter optimization process included the learning rate and weight decay parameters.
Gradient clipping was applied to mitigate potential gradient explosion problems and ensure stable learning, which is the usual practice for preventing the gradients from becoming too large, leading to the divergence during training~\cite{von2023transformers}.
The hyperparameter optimization used the \ac{WaB} platform's sweep functionality. 
The sweep was configured to utilize Bayesian optimization to maximize the F1 score on the validation set. 
As shown in Table~\ref{tab:gnnparameters}, the optimization explored various hyperparameters. 
This systematic exploration identified a combination of hyperparameters that best suit the data and the task at hand, resulting in a model capable of event localization.

\begin{table}[!t]
{\color{red}
\small
\vspace{-1mm}
\begin{center}
\caption{Hyperparameter search space.}
\vspace{-1mm}
\label{tab:gnnparameters}
\begin{tabular}{p{3.9cm} c}
\hline
\hfil \textbf{Parameter} & \textbf{Search space} \\
\hline
\multicolumn{2}{c}{\textbf{GNN hyperparameter search space}}  \\
\hline
Nr. Epochs (NE) & [20, 100] \\
Nr. Hidden Channels (HC) & \{16,\,32,\,64,\,128,\,256,\,512\} \\
Nr. HGT Attent. Heads (HAH) & \{1,\,2,\,4,\,8\} \\
Nr. HGT Layers (HL) & \{1,\,2,\,3,\,4\} \\
Nr. Initial Conv. Layers (ICL) & \{0,\,1,\,2,\,3,\,4\} \\
Nr. Final Conv. Layers (FCL) & \{0,\,1,\,2,\,3,\,4\} \\
Learning Rate (LR) & [0.00001, 0.01] \\
Weight Decay (WD) & [0.000001, 0.005] \\
Gradient Clipping Norm (MN) & [0.5, 5] \\
HGT Conv. Layer Type (CT) & [GAT, GCN] \\
\hline
\multicolumn{2}{c}{\textbf{DFS hyperparameter search space}} \\
\hline
Nr. Hidden Dims (HD) & \{16,\,32\,,64,\,128,\,256,\,512,\,1024\} \\
Nr. Hidden Layers (HL) & \{1,\,2,\,3,\,4\} \\
Learning Rate (LR) & [0.00001, 0.01] \\
Weight Decay (WD) & [0.000001, 0.005] \\
Gradient Clipping Norm (MN) & [0.5, 5] \\
\hline
\end{tabular}
\end{center}
\vspace{-1mm}}
\end{table}


\section{Evaluation Setup}
\label{sec:comparison}

For assessing the accuracy of the proposed analytical model of raw data for flow-guided localization \cd{(cf., Section~\ref{sec:prob_distrib})}, we will compare the raw data yielded by the model with the corresponding one generated by a simulator for objective performance benchmarking of flow-guided localization~\cite{lopez2023toward}. 
The simulator~\cite{lopez2023toward} features a significantly higher realism level than the model.
Specifically, it combines \ac{BVS}~\cite{geyer2018bloodvoyagers} for modeling the mobility of the nanodevices in the bloodstream and ns-3-based TeraSim~\cite{hossain2018terasim} for \ac{THz}-based nanoscale communication between the nanodevices and the outside world, accounting for the energy-related and other technological constraints (e.g., pulse-based modulation) of the nanodevices.

\ac{BVS} encompasses a comprehensive set of 94 vessels and organs, utilizing a coordinate system centered on the heart. 
All organs share an identical spatial depth, calibrated to a reference thickness of 4 cm, resembling the dimensions of a kidney. 
The simulator also assumes a predefined arrangement with arteries positioned anteriorly and veins posteriorly. 
The transitions between arteries and veins occur within organs, limbs, and head, which jointly account for 24 regions in the body, as indicated in Table~\ref{tab:body_parts}. 
The blood transitions from veins to arteries in the heart, signifying a shift from posterior to anterior flow.
The simulator models the flow rate based on the relationship between pressure differences and flow resistance.
This results in average blood speeds of 20~cm/sec in the aorta and 10~cm/sec in the arteries (\emph{Region type = 0}), and 2-4~cm/sec in veins (\emph{Region type = 1}). 
Transitions between arteries and veins are simplified with a constant velocity of 1~cm/sec (\emph{Region type = 2}).

TeraSim~\cite{hossain2018terasim} is the pioneering simulation platform tailored for modeling \ac{THz} (nano)communication networks. 
This platform accurately captures the unique capabilities of nanodevices and distinctive characteristics of THz signal propagation. 
TeraSim is integrated as a module within ns-3, a discrete-event network simulator, and it incorporates specialized physical and link layer solutions optimized for nanoscale THz communications. 
Specifically, at the physical layer, TeraSim implements pulse-based communication with omnidirectional antennas, catering to distances shorter than 1~m, assuming a single transmission window of nearly 10~THz. 
TeraSim incorporates two well-established protocols at the link layer, ALOHA and CSMA. 

\begin{figure*}[!t]
\centering
{\color{red}
\subfigure[THz-based in-body communication scenario (adapted from~\cite{gomez2023optimizing}).]{
\includegraphics[width=0.45\linewidth]{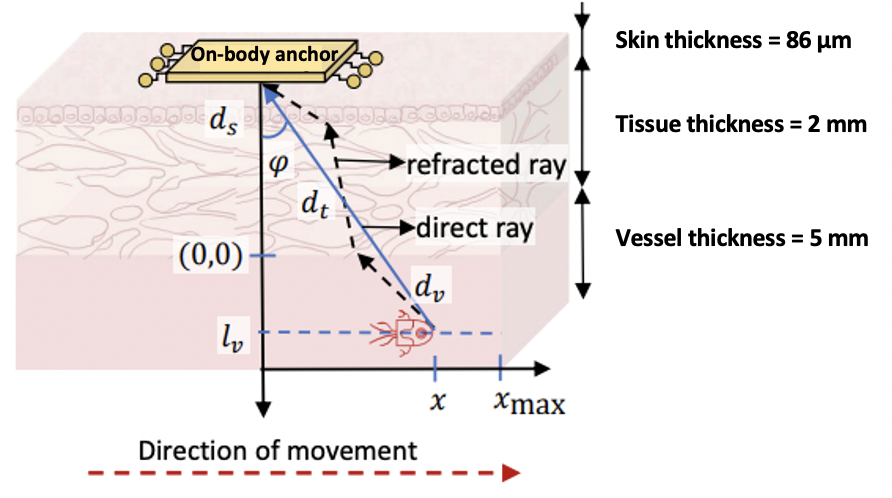}}
\subfigure[Indicative distances of successful transmission.]{
\includegraphics[width=0.33\linewidth]{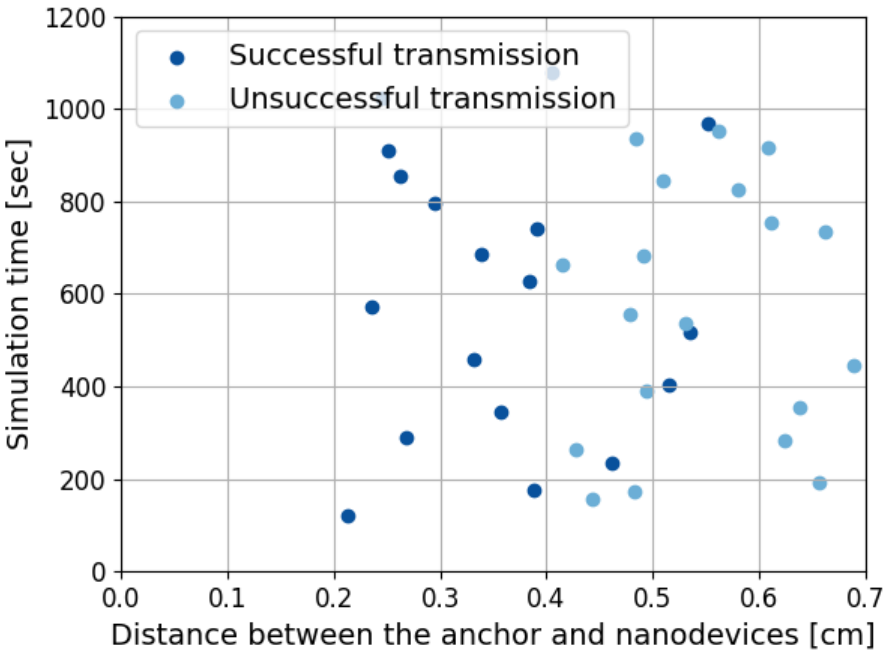}}
\vspace{-2mm}}
\caption{\textcolor{red}{modeling of THz-based in-body nanocommunication.}}
\label{fig:propagation}
\vspace{-3.5mm}
\end{figure*}

Additionally, we have introduced in TeraSim a shared \ac{THz} channel module that implements a frequency-selective channel simulating in-body wireless nanocommunication~\cite{gomez2023optimizing}, as shown in Figure~\ref{fig:propagation}a.
The simulator models the \ac{THz} channel by calculating the received power for each pair of communicating devices and scheduling the reception of packets based on corresponding propagation times. 
The model considers in-body path loss and Doppler effects by accounting for vessel, tissue, and skin thickness and attenuation, and analyzing changes in relative positions between nanodevices and the anchor over time.
The collision potential is modelled by calculating the \ac{SINR} and discarding the packet if the \ac{SINR} falls below a predefined reception threshold, known as the receiver sensitivity.

\begin{table}[!t]
    \centering
    \scriptsize
    \vspace{-1mm}
    \caption{BVS-based instantiation to 24 body regions.}
    \vspace{-3mm}
    \label{tab:body_parts}
    \begin{multicols}{3}
        \begin{tabular}{cl}
        \hline
        \textbf{ID} & \textbf{Body Part} \\
        \hline
        1 & Head \\
        2 & Thorax \\
        3 & Right shoulder \\
        4 & Left shoulder \\
        5 & Spleen \\
        6 & Right upper arm \\
        7 & Left upper arm \\
        8 & Liver \\
        \hline
        \end{tabular}        
        \begin{tabular}{cl}
        \hline
        \textbf{ID} & \textbf{Body Part} \\
        \hline
        9 & Right elbow \\    
        10 & Intestine \\
        11 & Right hand \\
        12 & Kidneys \\
        13 & Left elbow \\
        14 & Left hand \\
        15 & Right hip \\
        16 & Left hip \\
        \hline
        \end{tabular}
         \begin{tabular}{cl}
        \hline
        \textbf{ID} & \textbf{Body Part} \\
        \hline
        17 & Right knee \\     
        18 & Left pelvis \\
        19 & Left knee \\
        20 & Right pelvis \\
        21 & Right foot \\
        22 & Left foot \\
        23 & Lungs \\
        24 & Right heart \\
       \hline
        \end{tabular}
    \end{multicols}
\end{table}

{\color{red}
The anchor in the proximity of the heart, as well as the later discussed additional anchors, have been positioned in the communication range of the nanodevices, with communication distances as indicated in Figure~\ref{fig:propagation}b.
The practical implications of this assumption are discussed in Section~\ref{sec:discussion}.}
In the \ac{THz}-based communication between an anchor and nanodevices, the anchor transmits beacons at a constant frequency and power.
The nanodevices passively receive the beacons and actively send back responses, which consumes energy. 
The response packets from the nanodevices contain information about the time elapsed since their last passage through the heart and an event bit.
These data points are then used by a flow-guided localization approach to determine the location of an event. 
Whenever a nanodevice passes through the heart, the time elapsed since the last passage is reset to avoid accumulating multiple circulation periods. 
The event bit is set to ``1'' if a target event is successfully detected and resets in each passage through the heart.

Our analytical model has been instantiated on the 24 organ/limb regions, as modelled by the \ac{BVS}.
Other simulation parameters such as the transmit power, receiver sensitivity, number of nanodevices, operating frequency, and communication bandwidth have been consistently parameterized across the model and simulator. 
For each of the considered evaluation scenarios, two sets of raw data are generated, one with a varying probability of transmission $P_{trans}$ and ideal detection probability (i.e., $P_{det}=1$), and vice-versa (i.e., $P_{trans}=1$). 
These probabilities have been hard-coded in the simulator to assess the consistency of the raw data outputs between the model and the simulator.
A realistic scenario is also considered in which both probabilities are set to non-ideal values to assess the model's capabilities in capturing their joint effects on the raw data.

The following metrics are employed to assess the performance similarity across the raw datasets: 
\paragraph{\textbf{\ac{MW} test}} will be used to evaluate the similarity between the distributions of iteration times between the simulator and the model for body regions containing an event (i.e., for event bit $b=1$). 
\paragraph{\textbf{Square difference between \acp{ECDF}}} will be generated for event bit values $b=0$ for all body regions. 
The maximum vertical distance between \acp{ECDF} will be computed and averaged over all regions and provided graphically utilizing regular box plots, facilitating the qualitative interpretation of the data.
\paragraph{\textbf{\ac{KL} divergence}} will be employed to compare the difference between ratios of ones and zeros in each region for varying transmission and detection probabilities. 
This approach helps identify regions where the fraction of event bits $b=1$ is comparable across the simulator and the model.
Regions through which the nanodevices pass in each of their iterations through the bloodstream (lungs and right heart) are excluded from this metric.

We utilize the same simulator from~\cite{lopez2023toward} for assessing the performance of the proposed \ac{GNN}-based flow-guided localization approach \cd{described in Section \ref{sec:GNN_loc}}.
In the simulator, the nanodevices are assumed to have capacitors to store and \ac{ZnO} nanowires to collect energy. 
The charging of the capacitors is modelled as a \cd{logarithmic} process that takes into account the rate and interval of energy harvesting, as well as the storage capacity of the capacitors. 
The nanodevices exhibit intermittent behavior due to energy harvesting and storage constraints. 
This behavior is represented by a \emph{Turn ON} threshold, where a nanodevice turns on if its current energy level exceeds the threshold. 
Once the energy is depleted, the nanodevice turns off until its energy increases above the threshold.
When the nanodevices are turned on, they perform sensing tasks at a given frequency or granularity. 
Each task consumes a constant amount of energy, meaning that more frequent tasks require higher energy consumption at the nanodevice level. 
The event's location to be detected is assumed to be pre-programmed by the experimenter. 
A nanodevice is considered to detect an event if it is turned on and its location at the time of the sensing task execution is close to the event's location based on a predefined threshold (i.e., 1~cm).
The simulation parameters are summarized in Table~\ref{tab:paramaters}.

\section{Evaluation Results}
\label{sec:results}

\subsection{Analytical Modelling of Raw Data}

The \ac{MW} test results are aggregated across all regions and presented in Table~\ref{table:mannwhitney}. 
The results demonstrate that the null hypothesis of equally distributed datasets is predominantly accepted across regions, with a particularly strong performance observed for the ideal detection probability. 
Moreover, a higher variability pattern can be observed in the regions that do not meet the test's criteria, suggesting the potential influence of unspecified stochastic factors on the model's behavior in a small number of specific cases. 
Nonetheless, it is evident that the similarity between the model and the simulator distributions for event bit $b=1$ is notably high across the considered scenarios.

\begin{table}[!t]
\small
\vspace{-1mm}
\begin{center}
\caption{Simulation parameters.}
\label{tab:paramaters}
\vspace{-1mm}
\begin{tabular}{l r}
\hline
\textbf{Parameter} & \textbf{Value} \\
\hline
Generator voltage $V_g$ [V] & 0.42 \\
Energy consumed in pulse reception $E_{R_{X pulse}}$ [pJ] & 0.0 \\
Energy consumed in pulse transmission $E_{T_{X pulse}}$ [pJ] & 1.0 \\
Maximum energy storage capacity [pJ] & 800 \\
Turn ON/OFF thresholds [pJ] & 10/0 \\
Harvesting cycle duration [ms] & 20 \\
Harvested charge per cycle [pC] & 6 \\
Transmit power $P_{T_X}$ [dBm] & -20 \\
Operational bandwidth [GHz] & 10 \\ 
Receiver sensitivity [dBm] & -110 \\ 
Operational frequency [THz] & 1 \\ 
Simulation time [sec] & 1100 \\
Number of anchors & [1, 2, 3] \\
Number of nanodevices & 64 \\
Event sampling granularity [1/sec] & \hfil [1, 3, 5] \\
Event detection threshold [cm] & \hfil 1  \\
\hline
\end{tabular}
\end{center}
\vspace{-1.5mm}
\end{table}

\begin{figure}[!t]
\vspace{-1mm}
\centerline{\includegraphics[width=0.94\columnwidth]{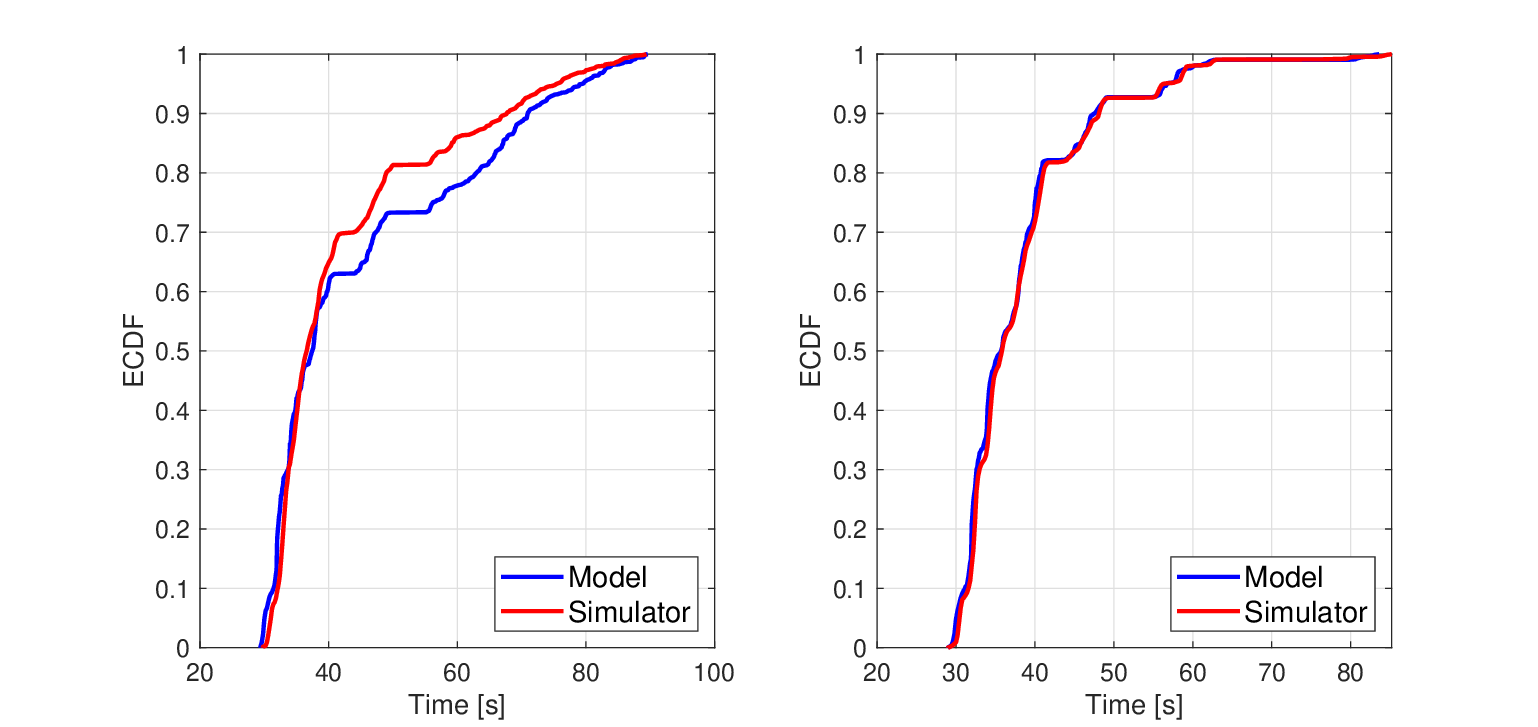}}
\caption{ECDF comparison between model and simulator for $P_{trans}=0.4$ (left) and $P_{det}=0.4$ (right) in thorax.}
\label{fig:ecdfp}
\vspace{-3mm}
\end{figure}

For scenarios involving event bit $b=0$, insights into the performance similarity between the model and the simulator are assessed utilizing \acp{ECDF}. 
The average maximum vertical distances between \acp{ECDF} are shown in Table~\ref{table:ecdf}. 
A higher level of similarity between ECDFs is observed for the transmission probability $P_{trans}=1$ compared to $P_{det}=1$. 
This difference can be attributed to the presence of compound iterations when varying the transmission probability (cf., Figure~\ref{fig:Probd}). 
It is worth noting that the discrepancy occurs within the distributions' midsection, while the tails are nearly indistinguishable. 
Figure~\ref{fig:ecdfp} depicts an illustrative instance of this behavior.
As visible, there are notable differences in a small number of specific cases, where the focal dissimilarity points are confined to specific parts of the distribution. Otherwise, the distributions feature significant similarity levels.

\begin{figure}[!t]
\centerline{\includegraphics[width=\columnwidth]{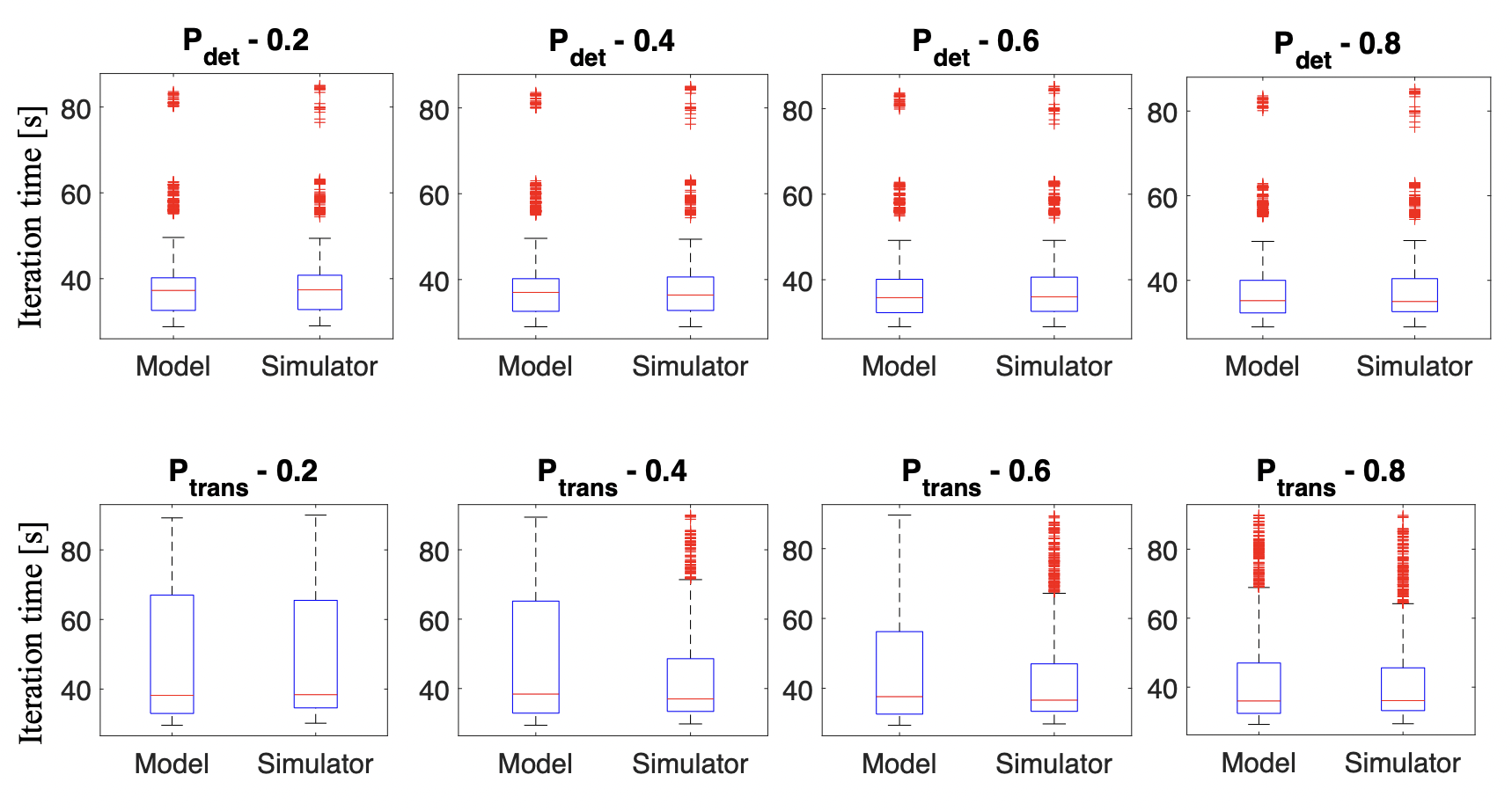}}
\caption{Iteration times with event bit $b=1$ for the head region.}
\label{fig:boxplots}
\end{figure}

To examine the distributions for event bit $b=0$, the distributions of iteration times in different scenarios for the head region are shown in Figure~\ref{fig:boxplots}.
As visible, the similarity in the data distributions is evident, apart in cases when $P_{trans}$ equals 0.4 and 0.6.  
Specifically, while the simulator-generated boxplots exhibit a substantial number of outliers, these outliers are considered within the third quartile in the model-related ones.
It is essential to recognize that although visually distinct, such outliers do not indicate a fundamental divergence in the dataset's characteristics. 
This can be checked for the case of transmission probability equaling 0.2, where the simulator-generated distribution is practically equal to the one originating from the model as the number of outliers decreases. 
These outliers are the root cause of the difference between \acp{ECDF}, as seen in \cd{Table}~\ref{table:ecdf}. 

\begin{table}[t]
  \centering
  \caption{Mann-Whitney test results for event bit $b=1$.}
    \label{table:mannwhitney}
  \begin{tabular}{cccc}
    \toprule
    \multirow{2}{*}{$P_{trans}$/$P_{det}$} & \multicolumn{2}{c}{Fraction of accepted Mann-Whitney tests} \\
    \cmidrule{2-3}
    & $P_{det}=1$ & $P_{trans}=1$ \\
    \midrule
    0.2 & 83\% & 92\% \\
    0.4 & 88\% & 79\% \\
    0.6 & 94\% & 75\% \\
    0.8 & 83\% & 79\% \\
    \bottomrule
  \end{tabular}
\end{table}

\begin{table}[t]
  \centering
  \caption{ECDF comparison for event bit $b=0$.}
  \label{table:ecdf}
  \begin{tabular}{cccc}
    \toprule
    \multirow{2}{*}{$P_{trans}$/$P_{det}$} & \multicolumn{2}{c}{Mean of vertical distance between ECDFs} \\
    \cmidrule{2-3}
    & $P_{det}=1$ & $P_{trans}=1$ \\
    \midrule
    0.2 & 0.11 & 0.081 \\
    0.4 & 0.12 & 0.084 \\
    0.6 & 0.1 & 0.083 \\
    0.8 & 0.1 & 0.08 \\
    \bottomrule
  \end{tabular}
\end{table}

\begin{figure}[t]
\vspace{-2mm}
\centerline{\includegraphics[width=\columnwidth]{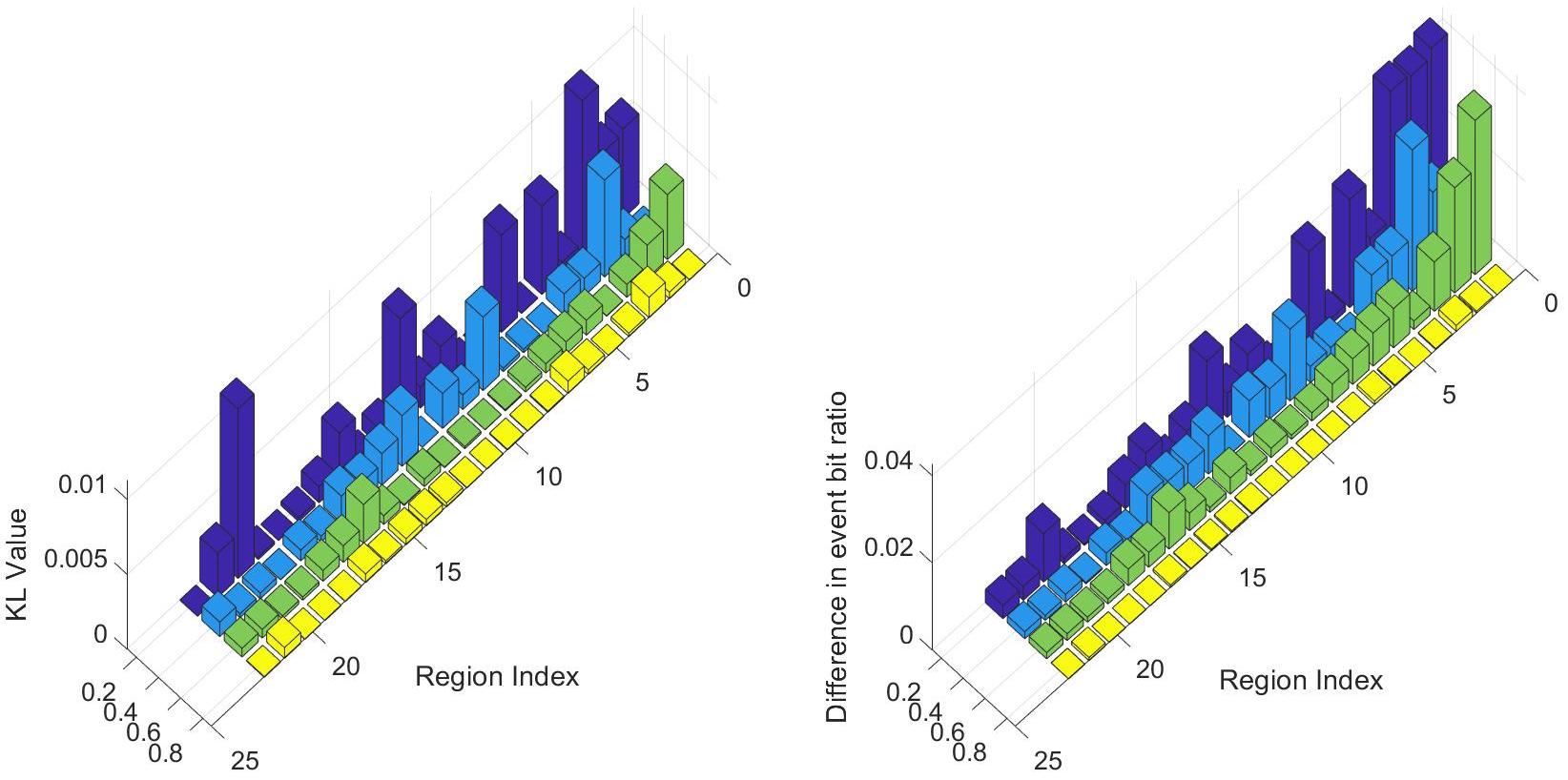}}
\caption{KL divergence and event bit ratio difference between the simulator and model with fixed detection probability $P_{det}$ and varying transmission probability $P_{trans}$. Region indices correspond to Table~\ref{tab:body_parts}.}
\vspace{-3mm}
\label{fig:metrics}
\end{figure}

The \ac{KL} divergence is used for assessing the similarity between the event bit $b=1$ ratio in the overall datasets obtained using the simulator and the model, with results depicted in Figure~\ref{fig:metrics}. 
As depicted, the arrangement of regions follows a descending order of $P_{trans}$ probability. 
The results demonstrate the increased in the similarity between ratios as a function of ascending transmission probability, in line with expectations. 
Notably, the peaks in the \ac{KL} divergence emerge within less probable regions coupled with reduced transmission probabilities. 
These peaks can be attributed to the scarcity of detection of events in these regions. 
Despite the marginal discrepancies in the ratios, their diminutive nature accentuates the \ac{KL} divergence, resulting in increased divergence in such cases. 
Notably, the observed differences are at most 0.04, underlining high similarity levels across scenarios.

\begin{figure}[t]
\centerline{\includegraphics[width=\columnwidth]{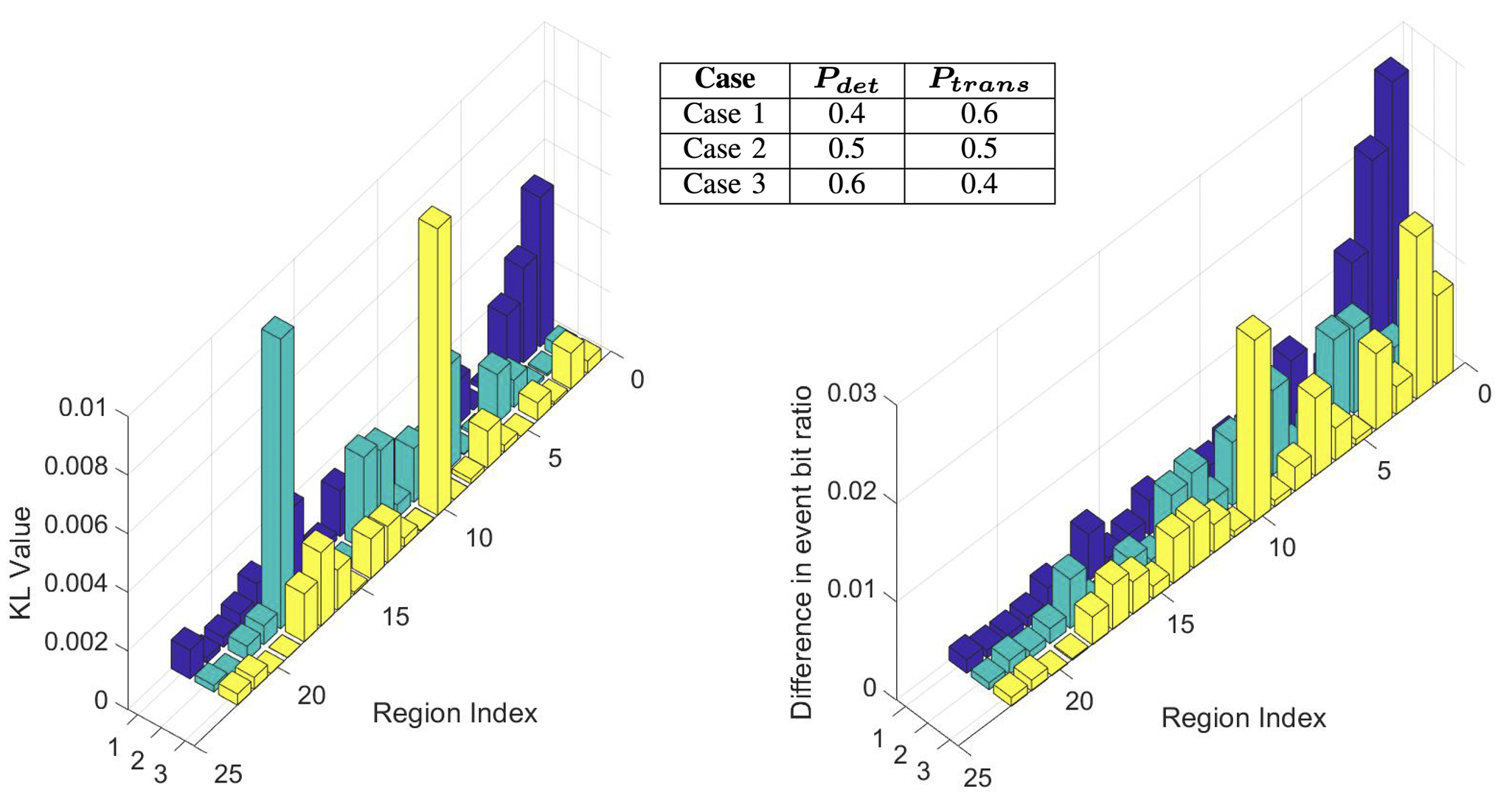}}
\caption{KL divergence and event bit ratio difference between the simulator and model in mixed parameter cases. Region indices correspond to Table~\ref{tab:body_parts}.}
\label{fig:klr}
\end{figure}

Lastly, several scenarios with mixed detection and transmission probabilities are studied to assess the model's accuracy for close-to-real-life cases.
Figure~\ref{fig:klr} depicts the outcomes for three distinct cases, while the depicted results affirm comparable performance of the model and the simulator in realistic scenarios. 
However, the trends within the \ac{KL} divergence are not entirely comparable due to a few outliers in the simulator-generated raw data, indicating the presence of smaller stochastic factors beyond the model's scope. 

\begin{figure*}[!t]
\centering
\vspace{-3mm}
\subfigure[Main organs, limbs, and head]{
\includegraphics[width=0.45\linewidth]{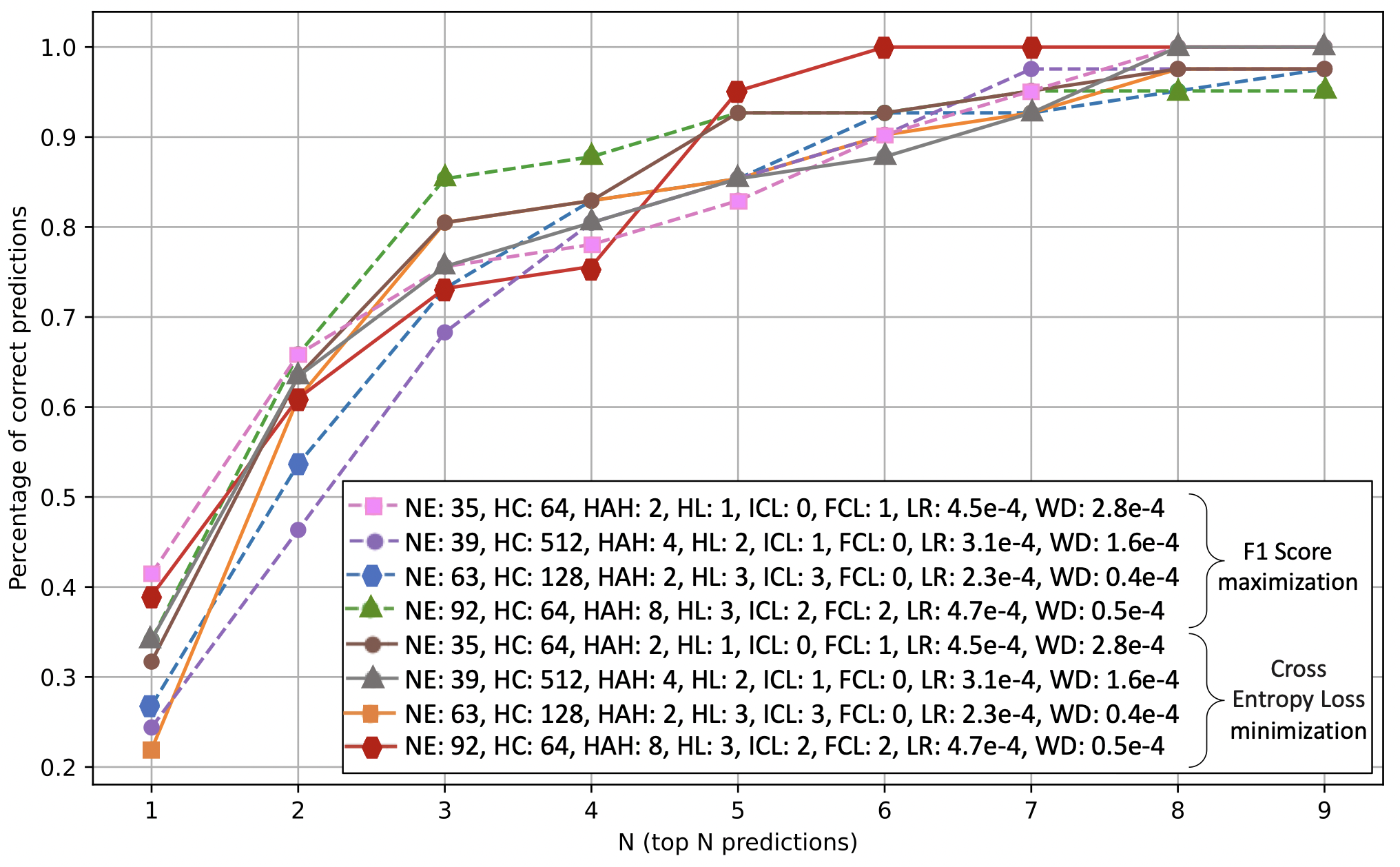}}
\subfigure[Entire bloodstream]{
\includegraphics[width=0.45\linewidth]{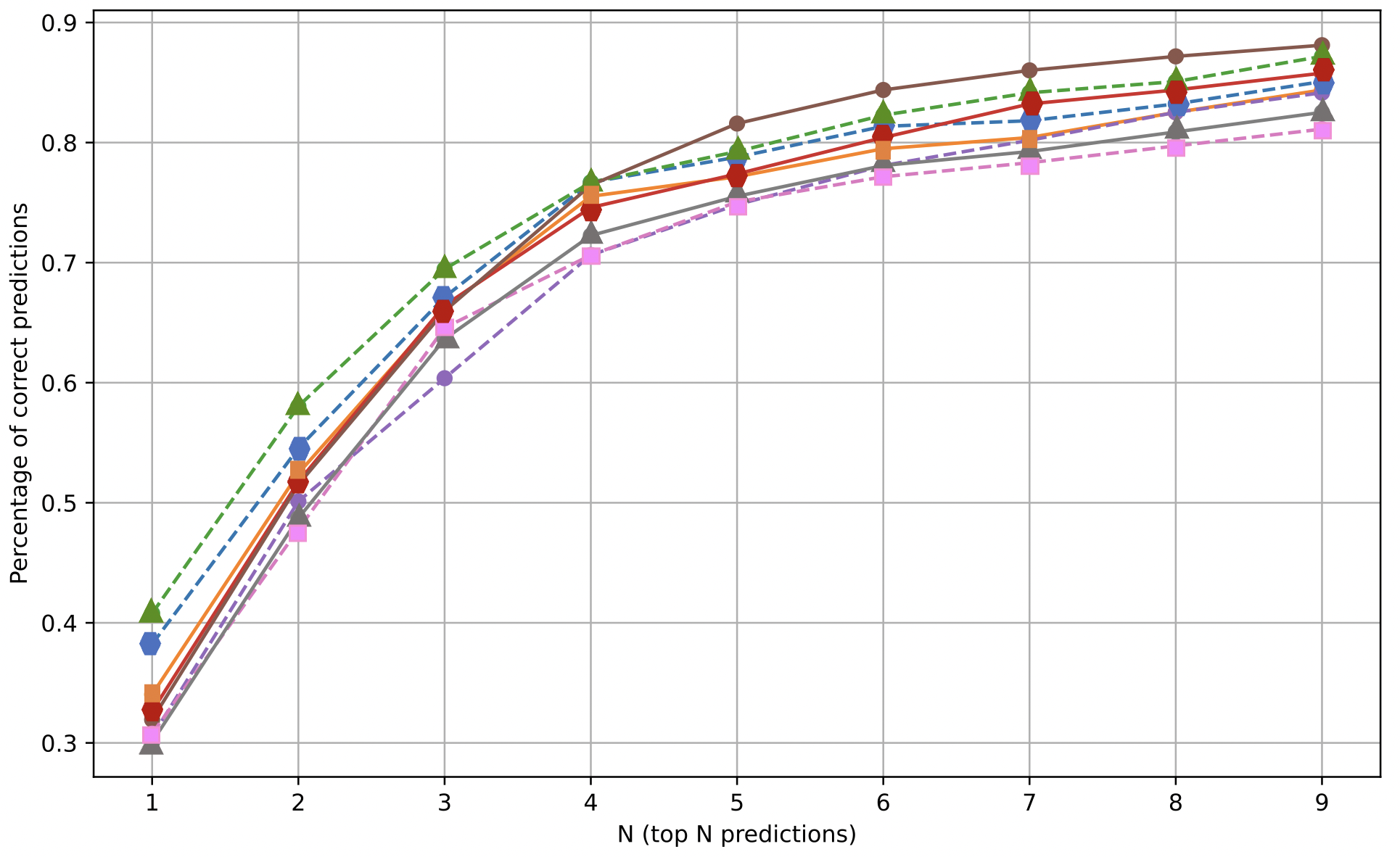}}
\vspace{-1.5mm}
\caption{Hyperparameter tuning (see Table~\ref{tab:gnnparameters} for understanding the legend).}
\label{fig:hyperparameters}
\vspace{-4mm}
\end{figure*}

\begin{figure}[t]
\vspace{-1mm}
\centering
 \includegraphics[width=0.98\linewidth]{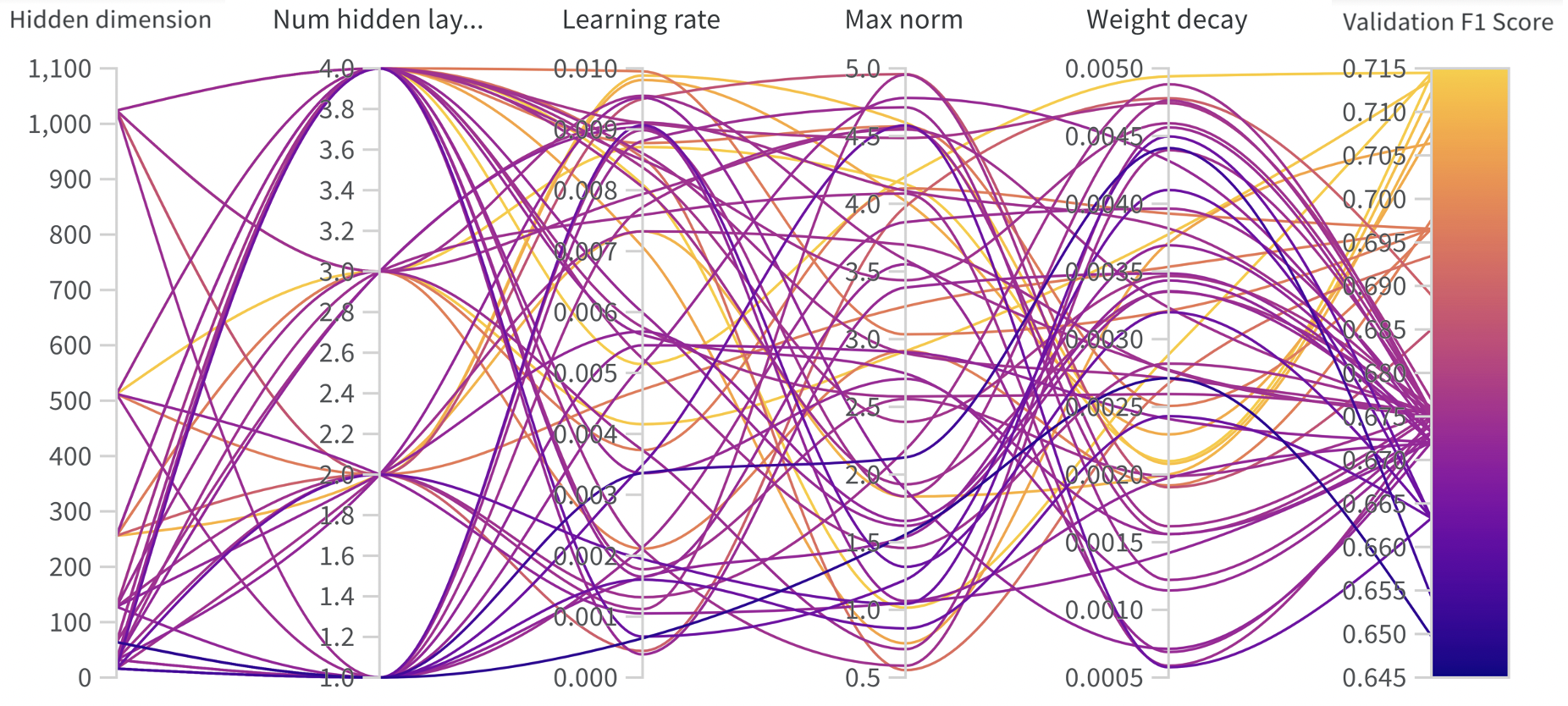}
\caption{Hyperparameter tuning for the additional anchors.}
\label{fig:hyperparameter_additional}
\end{figure}

\subsection{GNN-based Flow-guided Localization}
The hyperparameter tuning of the proposed \ac{GNN} model is depicted in Figure~\ref{fig:hyperparameters}, which includes the cases in which the events might be located solely in limbs, organs, and head, i.e., the regions in which the blood speeds are low, as well as accounting for the entire bloodstream.
{\color{red}Additionally, Figure~\ref{fig:hyperparameter_additional} depicts the hyperparameter tuning of the DFS-based extension to multi-anchor systems.  
Note that for both the GNN model and DFS-based extension, only the best-performing subset of the considered hyperparameters is depicted for clarity, and our depiction includes optimizing the F1 score and \ac{CE} loss.}

We first consider a single-anchor system that is, by design, unable to distinguish between left and right body sides due to comparable circulation times of the corresponding regions. 
Hence, as the hyperparameter tuning objective, we consider multiple targets as correct, as depicted on the x-axis in the graphs.
This approach also provides an indication of how many body regions are, on average, incorrectly classified. 
Considering the best-performing hyperparameter set as shown in Figure~\ref{fig:hyperparameters}a), the correct region is always within the 6 most likely regions the network outputs. 
This provides a primer for constraining diagnostic searches, exploratory surgeries, and similar medical procedures. 
In the remainder of our evaluation, we utilize the model indicated with green line in Figure~\ref{fig:hyperparameters} \cd{(i.e., NE: 92, HC: 64, HAH: 8, HL: 3, ICL: 2, FCL: 2, LR: 4.7e-4, WD: 0.5e-4)} due to its close-to-optimal performance in both scenarios for the number of correct targets equaling 1. 

\begin{figure}[t]
\centering
 \includegraphics[width=\linewidth]{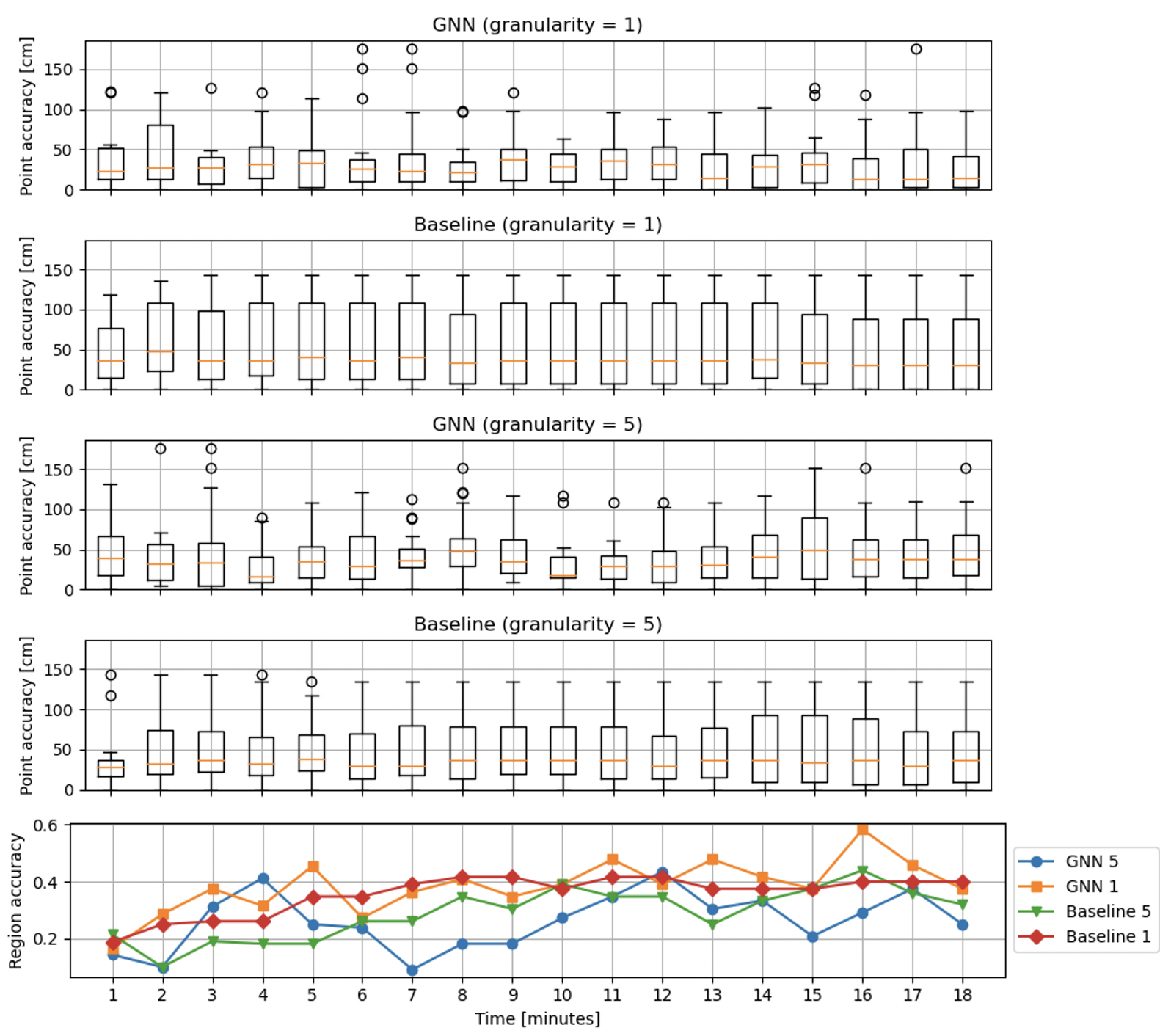}
\vspace{-3mm}
\caption{Comparison with the current State of the Art.}
\label{fig:gnn_vs_baseline}
\end{figure}

In Figure~\ref{fig:gnn_vs_baseline}, we compare the performance of the \ac{GNN} model with the \ac{SotA} baseline from~\cite{gomez2022nanosensor,lemic2023insights}.
The baseline is a \ac{NN} solution that implements three fully connected layers, with PReLU activation function for the first two and log-softmax for the last. 
The first two layers feature a dropout for regularization and batch normalization for stabilizing the learning process. 
The hidden layer's size is 512, and the model is trained to classify 25 classes. 
The approach utilizes the Negative Log Likelihood loss due to its ability to handle unbalanced datasets, as well as the Adam optimizer due to its dynamic learning rate adaptation and its ability to operate with relatively simple fine-tuning of the hyperparameters.

{\color{red}
Based on the discussion from Section~\ref{sec:related_work}, our results are depicted as a function of the duration of the administration of the nanodevices in the bloodstream, as indicated on the x-axis in Figure~\ref{fig:gnn_vs_baseline}.
Consistent training and evaluation raw datasets have been utilized across the considered approaches.
The runtime performance obtained through an 18-minute-long runtime dataset is depicted, with both models trained on a 30-minute-long one.  
In our results, we consider two different event sampling granularities of 1 and 5 samples per second. 
We do that to demonstrate the complex interplay between energy consumption increase due to more frequency event sampling, which increases the probability of detecting such events while simultaneously degrading the wake-up duration of the nanodevices.
}

The comparison is carried out along a set of heterogeneous performance metrics characterizing the accuracy of flow-guided localization. 
Specifically, the point accuracy metric indicates the amplitude of localization errors. It is derived as the Euclidean distance between the true location of the event and the estimated one, with the estimated one being modeled as the centroid of the estimated region.
The point accuracy results are depicted in the regular boxplot fashion, indicating the distribution of such errors for 24 randomly located events at each time-step indicated on the x-axis in Figure~\ref{fig:gnn_vs_baseline}, one in each of the 24 body regions with blood speeds of 1~cm/sec, as modeled by the BloodVoyagerS.
The region accuracy is defined as the percentage of correctly estimated regions.
The figure shows that the proposed approach outperforms the baseline regarding point accuracy, regardless of the considered granularity.
For example, 18 minutes after deploying either solution and considering the sensing granularity of 1 sample per second, the point accuracy distribution of the \ac{GNN}-based approach is bounded to less than 100~cm of error. At the same time, in the baseline, a significant number of estimates feature errors bounded by 150~cm, representing an improvement of more than 30\% over the baseline.

Figure~\ref{fig:gnn_vs_baseline} also illustrates that, as the runtime duration increases, the point and region estimation accuracies exhibit only slight to no improvement, regardless of the solution under consideration.
For instance, the region estimation accuracy increases by approximately 20 to 25\%. 
This behavior can be attributed to two key factors that significantly impact the performance of the considered solutions.
The first one is the general principle that \ac{ML} models' performance is enhanced when provided with a larger volume of raw data for making predictions. 
However, it is crucial to simultaneously consider the challenges associated with \ac{THz} communication between the nanodevices and the anchor. 
These challenges include high in-body attenuation, the nanodevices' high mobility, and self-interference between different nanodevices attempting to communicate with the anchor simultaneously.
Due to these obstacles, the communication becomes unreliable, resulting in the anchor often not receiving raw data from certain nanodevices at specific time points.
More problematically, in such cases the nanodevices do not reset their iteration times and event bits.  
Consequently, when the data is eventually reported to the anchor, the reported iteration times represent a combination of multiple iterations, while the event bit may be erroneous. 
In other words, the event was detected in one of the iterations but propagated through several iterations, some of which did not actually feature the event. 

{\color{red}
Finally, one should adequately dimension the event sampling granularity due to its complex relationship with the operation of flow-guided localization. 
This can be best observed from the performance results of the baseline approach in Figure~\ref{fig:gnn_vs_baseline}.
As visible, the selected sampling granularity plays a significant role in the performance of the baseline, with the sampling granularity of one sample per second yielding significantly better region and point estimation accuracies than the one sampled at five samples per second.
}

Moreover, the intermittent operation of nanodevices, driven by energy harvesting, can lead to situations where a nanodevice misses detecting an event because it was turned off, even though it passed through a region where the event was located. 
This behavior indicates that, although increasing the amount of data input into the models should enhance the accuracy of estimation, the highly erroneous nature of the data counterbalances these improvements, resulting in a ``flat'' performance in terms of region detection and point accuracies for both solutions under consideration.
In other words, the results suggest that none of the considered approaches can, to the full extent, deal with the erroneous nature and complexity of the raw data.
Hence, improvements primarily along the lines of introducing additional anchors will be required to optimize the accuracy of flow-guided localization.

\begin{figure}[!t]
\vspace{-1mm}
\centering
\includegraphics[width=\linewidth]{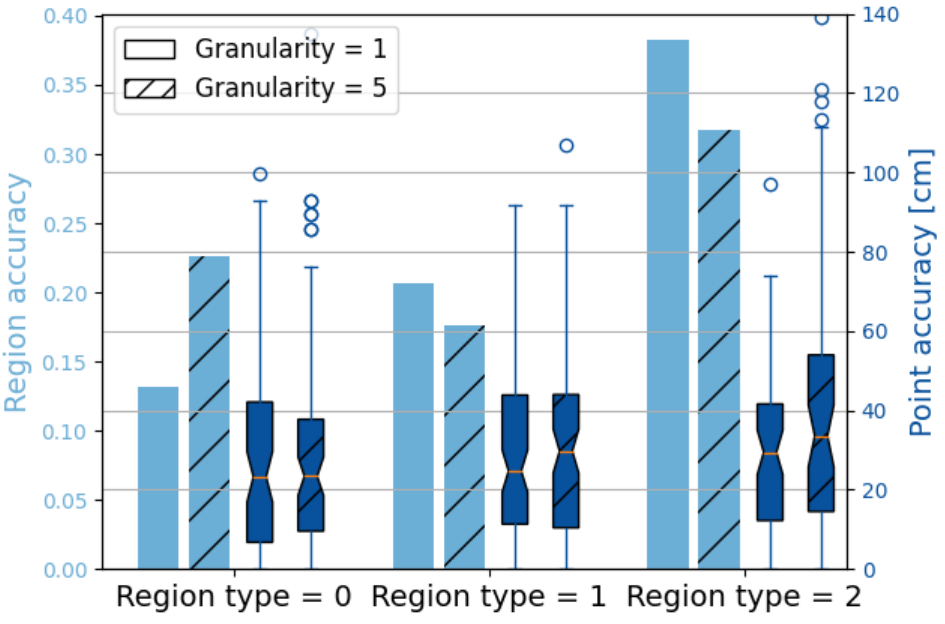}
\caption{Coverage of GNN-based flow-guided localization.}
\label{fig:coverage}
\vspace{-2.8mm}
\end{figure}

Apart from the enhancement in the point accuracy, the proposed \ac{GNN}-based approach is, by design, able to classify different regions throughout the bloodstream, including the ones in which the blood speeds are higher than 1~cm/sec, i.e., \emph{Region type = \{0, 1\}} in Figure~\ref{fig:coverage}.
This is in contrast to the baseline, which yields meaningless (i.e.,~0\%) region classification accuracy for these region types.
As visible in the figure, the \ac{GNN} approach can still maintain meaningful accuracy levels for the regions where the blood is faster than 1~cm/sec.
For example, in the region with the slowest blood speeds, the \ac{GNN}-based approach achieves the classification accuracy of almost 40\%, which reduces in regions with increased blood speeds.
The results also indicate that the event sampling granularity, i.e., the frequency of sensing that a nanodevice performs, significantly affects the accuracy of the proposed approach.
Specifically, our results indicate that sampling should happen more frequently in regions with fast blood speeds to avoid missing events. In contrast, the sampling frequency should be reduced for accuracy optimization when the blood is slower.

\begin{figure}[!t]
\vspace{-1mm}
\centering
\includegraphics[width=\linewidth]{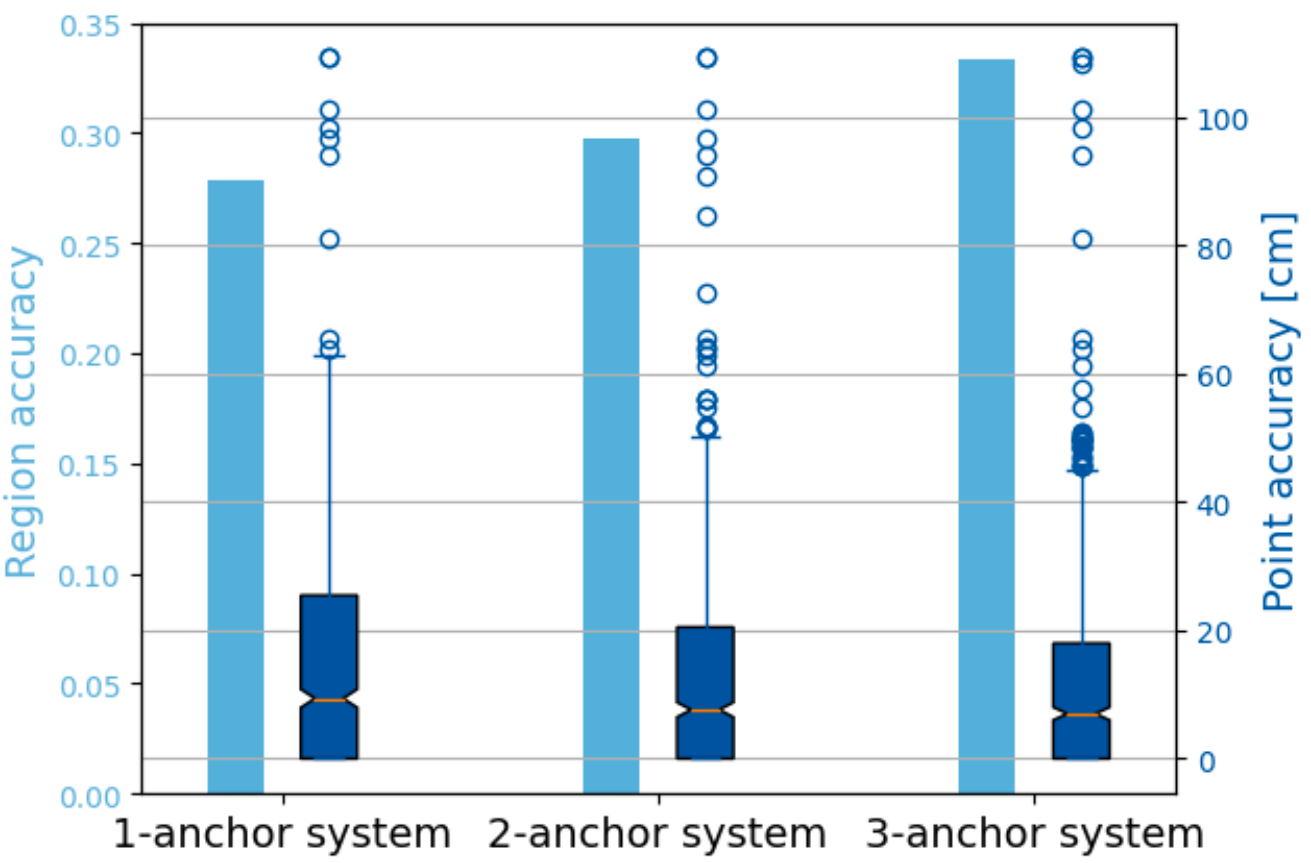}
\caption{{\color{red}Performance of example extensions to multianchor systems.}}
\label{fig:multianchor}
\end{figure}

{\color{red}
As mentioned, multi-anchor systems will be advantageous in dealing with the complexity and erroneous nature of the raw data.
Example extensions to two- and three-anchor systems are depicted in Figure~\ref{fig:multianchor} for the GNN model trained and evaluated on the entire bloodstream (n.b., all 94 BVS-modeled regions).
The second anchor is strategically placed at the axillary vein (Latin name: \textit{vena axiliaris sinister}) in the vicinity of the left armpit, as at this location it can communicate with the nanodevices passing through all cardiovascular paths of that arm.  
In the three-anchor system, the third anchor was placed at the symmetric location on the right arm, within the communication range with the nanodevices passing through that arm.
As visible from the figure, the multi-anchor extensions improve the performance of flow-guided localization across both metrics. 
Note that these results have been generated by strategically positioning the additional anchors, yet without searching for their optimal locations, which is considered beyond the scope of this work.  
Nonetheless, we expect optimal positioning of additional anchors to further refine flow-guided localization performance. 
Finally, introducing even more anchors, primarily on the legs, will intuitively also substantially benefit the performance of flow-guided localization.
}

{\color{red}
\section{Discussion, Challenges, and Future Research Directions}
\label{sec:discussion}

In the following, we discuss selected challenges and promising research directions on in-body nanoscale communication, system-level concerns, and data engineering and scientific challenges of eventually deploying GNN-based flow-guided nanoscale localization.  

\subsection{In-body Nanoscale Communication}

As mentioned, the on-body anchors considered in this work have been strategically positioned in the nanodevices' communication range of less than 0.5~cm. 
Such strategic positioning might be feasible for specific locations on the body (e.g., wrists) and only for some patients, with the main criteria being their body mass index. 
However, generally such positioning cannot be achieved noninvasively with the currently achievable range of \ac{THz}-based in-body communication. 
Ideally, future research in this domain should aim toward enhancing the communication distances between on-body anchors and in-body nanodevices so that the anchors can eventually be positioned on the body in a non-invasive manner. 

Moreover, note that the proposed flow-guided localization system does not require the anchors to communicate with the nanodevices but solely to indicate their vicinity.  
If this indication is received by a nanodevice, the nanodevice will attempt to report its raw data to the anchor, regardless of the anchor's role (i.e., anchor in the vicinity of the heart vs. supporting/additional anchors).
Nonetheless, if the anchor located in the vicinity of the heart indicated its vicinity, the nanodevice should additionally reset its raw data indicators (i.e., event detection indicator and circulation time).

The authors in~\cite{jornet2012joint,lemic2022toward} envision the utilization of ultrasound power transfer between on-body anchors and in-body nanodevices.
Such an approach might be used as a wake-up radio for the nanodevices, upon which the nanodevices would report their raw data to the anchors.
In such a scenario, the nanodevices' reception sensitivity, which cannot be very high due to their form factor, becomes irrelevant.
The relevance then shifts to enhancing the receiver sensitivity of the anchors, which features higher potential for future improvements due to their macro-scale.  
It is also worth mentioning that the development of ultrasound-based wake-up receivers is an active area of research, with prototypical implementations approaching the scales required for cardiovascular nanodevices~\cite{simeoni2020long}. 
Similarly, on-body devices that can focus ultrasound waves toward a target location deeply in the body are within reach, e.g.,~\cite{chen2018deep}.

If the enhancements in the receiver sensitivity of the on-body anchors and utilization of ultrasound for advertising the vicinity of the anchors and "blasting" the energy-harvesting nanodevices with the energy required for transmission is still insufficient for communicating the raw data to the outside world in an unobtrusive manner, we see two main alternative avenues.
The first one is to possibly replace \ac{THz}-based communication with alternative communication paradigms with extended range. 
Here, one might consider fat-based electromagnetic communication~\cite{engstrand2022end} as the prime candidate for the scenario at hand due to its potential for miniaturization~\cite{vizziello2023intra} and low latency communication (n.b., in contrast to molecular communication).
Finally, communication might also be enabled by utilizing long-term small-scale implants at strategic locations in the body~\cite{khaleghi2018radio}.
There, they would be supporting short-range THz-communication with the in-body nanodevices, as well as act as relays toward the outside world through, e.g., fat-based wireless communication~\cite{engstrand2022end}, low-frequency RF-backscattering~\cite{khaleghi2018radio}, or through wiring. 

\subsection{System Design Space Exploration}

One limitation of our work comes from the mobility of the in-body nanodevices, which was substantially simplified compared to what is expected in the eventual deployments of the system.
Specifically, the human bloodstream is a significantly more complex system than the BVS model, and future improvements are needed in terms of more precise modeling of the pathways the nanodevices might take, as well as more accurately capturing the peculiarities of the blood flow by, e.g., better accounting for its vortex and laminar nature. 

Despite the simplifications in the nanodevices' mobility, the accuracy of the proposed flow-guided localization approach is still modest.
Nonetheless, the space for improvement from the system-level perspective is abundant.  
One promising direction for such improvement lies in more optimal positioning and dimensioning of the supporting on-body anchors.
As an example, the dimensioning might intuitively benefit the localization performance if additional anchors are placed in the vicinity of the common iliac veins near the hips (Latin name: \textit{vena iliaca communis}) for distinguishing between the left and right counterparts in the leg regions.
Moreover, the circulation time might not be a suitable indicator of a region within the body core due to the small variability of such measurements across core regions.
In such a scenario, multi-anchor flow-guided localization might be combined with proteome fingerprinting of the core body~\cite{wendt2023proteome}.   

In more general terms, more research is needed on assessing the effects of different system parameters, such as the number of nanodevices, sampling granularity, or longer duration of the nanodevices' administration in the bloodstream, to further optimize the proposed approach's performance.
In addition, dynamic adaptation of certain system parameters based on the context of their operation might be a feasible option for further performance enhancements.
For example, the sampling granularity might be dynamically adapted to the blood speeds, thereby maximizing the localization accuracy while minimizing the energy consumption at the nanodevice level.
Nonetheless, the fact that meaningful accuracy has been observed for the entire bloodstream indicates that the \acp{GNN} will eventually be able to reach high region classification accuracy and full body coverage. 

Still, it should be noted that if the proposed approach cannot eventually yield high localization accuracy as the complexity of the nanodevices' mobility is enhanced to meet the intricacies of realistic bloodstreams, this does not necessarily imply the infeasibility of flow-guided localization.
In such a case, one should consider reducing the region granularity of flow-guided localization, given that a less granular region indication would still be beneficial from the diagnostic perspective.
In addition, one might consider a different set of applications with less strict requirements.
For example, the nanodevices' utilization for full body-profiling of biological parameters such as temperature or oxygenation would still feature an immerse diagnostic value (e.g.,~\cite{geurts2016temporal,zhang2015dna}), mainly if realizable with non-invasive anchor deployment.
Finally, the nanodevices might be used in lower quantities for assessing the structural health of the bloodstream, where the objective would be to detect anomalies in their mobility, in contrast to sensing events and supporting their localization.
Note that a structural health assessment of the bloodstream might be required before administering more nanodevices for other diagnostic purposes to avoid issues such as clotting.

\subsection{Data Manipulation and Processing}

Flow-guided localization is envisioned to be deployed in the bloodstreams of different individuals with varying biologies.
Such localization is usually based on \ac{ML}, requiring significant training and corresponding data. 
This training data is hard to obtain individually. However, the tunable nature of the proposed analytical model could be used for capturing the differences in the raw data across bloodstreams.
We consider evaluating this aspect of the model as a promising direction for future research, where we envision its utilization comparably to administering anesthesia, i.e., based on physiological indicators such as age, sex, height, and weight. 
Evaluating this aspect of the system will require modeling the differences between individual biologies, where \ac{ML} tools could be applied, as shown in~\cite{gomez2023fine}.

For an individual patient, temporal differences in the raw data stream for flow-guided localization are expected due to performed activities, biological conditions (e.g., diseases), and environmental changes that the individual experiences (e.g., temperature, humidity).
These will change the raw input data stream for flow-guided localization, in turn degrading its performance. 
We consider the adaptation of the model based on~\cite{gomez2023fine} as a promising future research direction, as it would enable it to capture these slight changes in individual bloodstreams, which we envisage to be used for adapting flow-guided localization based on physiological indicators such as blood pressure or heart rate.

Future work could consider assessing and enhancing the adaptability of the \ac{GNN} model on varying environmental conditions.
In that regard, the model could be extended with dynamic spatial modelling~\cite{wu2020comprehensive}, which is envisioned to enable it to operate well for a changing number of anchors, supporting scenarios in which the users wear multiple anchors, but eventually take some off (e.g., the ones on wrists) without the need for excessive retraining. 
The model's aspect of relating the nanodevice’s circulation time with the circuit length and velocity could be extended with dynamic temporal modelling~\cite{wu2020comprehensive}. 
Dynamic temporal modeling is envisaged to make the model "body agnostic," i.e., by design adaptable to different physical conditions such as an accelerated pulse during exercise, adaptable region classification granularities, and diverse bloodstreams, potentially without retraining.  

Finally, one interesting direction for improvement is to introduce heterogeneity~\cite{zhang2019heterogeneous} in the \ac{GNN} model in the sense of dynamically adjusting the number of output variables based on the features of input data.
By doing so, we hypothesize it may become capable of independently estimating the locations of multiple events simultaneously, paving the way toward its eventual deployment in the bloodstream. 
We envision this approach to also provide an added layer of interpretability for the model’s predictions~\cite{ying2019gnnexplainer}.
}

{\color{red}
\section{Conclusions}
\label{sec:conclusion}

In this work, we have proposed a\cd{n} analytical model of raw data for flow-guided localization.
Based on the nanodevices' communication and energy-related capabilities, the model outputs iteration times and event detection indicators that existing flow-guided localization approaches can use for localizing biological events. 
The model's output was compared with the equally parameterized one from a simulator for flow-guided localization with a higher level of realism, indicating a significant level of similarity.
Moreover, we have proposed a \acf{GNN} model for \acf{THz}-supported flow-guided nanoscale localization, as well as its \acf{DFS}-based extension to multi-anchor localization systems.
We have shown that the model outperforms the existing \acf{SotA} approaches in terms of localization accuracy, extends the coverage of such localization to the entire bloodstream, and supports anchor extensions for accuracy refinements.
We consider flow-guided localization a promising application of in-body nanonetworks in future precision medicine due to the numerous health benefits it is expected to bring.
This work serves as an invite for the community to further advance this nascent yet fast-developing research field\footnote{The code, data, and simulator from~\cite{lopez2023toward} are available for reproducibility:\\\url{https://bitbucket.org/filip_lemic/flow-guided-localization-in-ns3/src/parallel/}}.
}


\vspace{-1mm}
\section*{Acknowledgements}
This work was supported by the European Union’s Horizon Europe programme (grant nº 101139161 — INSTINCT project) and the Spanish Ministry of Economic Affairs with EU—NextGeneration EU funding, under the PRTR (Call UNICO I+D 5G 2021, Grant TSI-063000-2021-6-Open6G).

\vspace{-1mm}
\renewcommand{\bibfont}{\footnotesize}
\printbibliography

@article{gomez2022nanosensor,
  title={Nanosensor location estimation in the human circulatory system using machine learning},
  author={G{\'o}mez, Jorge Torres and Kuestner, Anke and Simonjan, Jennifer and others},
  journal={IEEE Transactions on Nanotechnology},
  volume={21},
  pages={663--673},
  year={2022},
  publisher={IEEE}
}

@article{lopez2023toward,
  title={Toward Standardized Performance Evaluation of Flow-guided Nanoscale Localization},
  author={L{\'o}pez, Arnau Brosa and Lemic, Filip and Struye, Jakob and others},
  journal={arXiv preprint arXiv:2303.07804},
  year={2023}
}

@article{velickovic2017graph,
  title={Graph attention networks},
  author={Velickovic, Petar and Cucurull, Guillem and Casanova, Arantxa and others},
  journal={STAT},
  volume={1050},
  number={20},
  pages={10--48550},
  year={2017}
}

@article{jornet2012joint,
  title={Joint energy harvesting and communication analysis for perpetual wireless nanosensor networks in the terahertz band},
  author={Jornet, Josep Miquel and Akyildiz, Ian F},
  journal={IEEE Transactions on Nanotechnology},
  volume={11},
  number={3},
  pages={570--580},
  year={2012},
  publisher={IEEE}
}

@article{jornet2014femtosecond,
  title={Femtosecond-long pulse-based modulation for terahertz band communication in nanonetworks},
  author={Jornet, Josep Miquel and Akyildiz, Ian F},
  journal={IEEE Transactions on Communications},
  volume={62},
  number={5},
  pages={1742--1754},
  year={2014},
  publisher={IEEE}
}

@article{emerich2003nanotechnology,
  title={Nanotechnology and medicine},
  author={Emerich, Dwaine and Thanos, Christopher},
  journal={Expert opinion on biological therapy},
  volume={3},
  number={4},
  pages={655--663},
  year={2003},
  publisher={Taylor \& Francis}
}

@article{caso2019vifi,
  title={ViFi: Virtual fingerprinting WiFi-based indoor positioning via multi-wall multi-floor propagation model},
  author={Caso, Giuseppe and De Nardis, Luca and others},
  journal={IEEE Transactions on Mobile Computing},
  volume={19},
  number={6},
  pages={1478--1491},
  year={2019},
  publisher={IEEE}
}

@article{senturk2022internet,
  title={Internet of Nano, Bio-Nano, Biodegradable and Ingestible Things: A Survey},
  author={Senturk, Seyda and Kok, Ibrahim and Senturk, Fatmana},
  journal={arXiv preprint arXiv:2202.12409},
  year={2022}
}

@article{muzykantov2011targeted,
  title={Targeted therapeutics and nanodevices for vascular drug delivery: quo vadis?},
  author={Muzykantov, Vladimir R},
  journal={IUBMB life},
  volume={63},
  number={8},
  pages={583--585},
  year={2011},
  publisher={Wiley Online Library}
}

@article{gao2007nanowire,
  title={Nanowire piezoelectric nanogenerators on plastic substrates as flexible power sources for nanodevices},
  author={Gao, Pu Xian and Song, Jinhui and Liu, Jin and Wang, Zhong Lin},
  journal={Advanced Materials},
  volume={19},
  number={1},
  pages={67--72},
  year={2007},
  publisher={Wiley Online Library}
}

@article{jornet2013graphene,
  title={Graphene-based plasmonic nano-antenna for terahertz band communication in nanonetworks},
  author={Jornet, Josep Miquel and Akyildiz, Ian F},
  journal={IEEE Journal on selected areas in communications},
  volume={31},
  number={12},
  pages={685--694},
  year={2013},
  publisher={IEEE}
}

@article{akyildiz2008nanonetworks,
  title={Nanonetworks: A new communication paradigm},
  author={Akyildiz, Ian F and Brunetti, Fernando and Bl{\'a}zquez, Cristina},
  journal={Computer Networks},
  volume={52},
  number={12},
  pages={2260--2279},
  year={2008},
  publisher={Elsevier}
}

@article{ying2019gnnexplainer,
  title={Gnnexplainer: Generating explanations for graph neural networks},
  author={Ying, Zhitao and Bourgeois, Dylan and You, Jiaxuan and Zitnik, Marinka and Leskovec, Jure},
  journal={Advances in neural information processing systems},
  volume={32},
  year={2019}
}

@inproceedings{gomez2023fine,
  title={Fine-tuned Circuit Representation of Human Vessels through Reinforcement Learning: A Novel Digital Twin Approach for Hemodynamics},
  author={Gomez, Jorge Torres and Spicher, Nicolai and Rios, Jorge Luis Gonz{\'a}lez and Dressler, Falko},
  booktitle={ACM International Conference on Nanoscale Computing and Communication},
  pages={46--52},
  year={2023}
}

@inproceedings{moayeri2016perfloc,
  title={PerfLoc (Part 1): An extensive data repository for development of smartphone indoor localization apps},
  author={Moayeri, Nader and Ergin, Mustafa Onur and others},
  booktitle={Annual International Symposium on Personal, Indoor, and Mobile Radio Communications},
  pages={1--7},
  year={2016},
  organization={IEEE}
}

@article{hossain2018terasim,
  title={TeraSim: An ns-3 extension to simulate Terahertz-band communication networks},
  author={Hossain, Zahed and Xia, Qing and Jornet, Josep Miquel},
  journal={Nano Communication Networks},
  volume={17},
  pages={36--44},
  year={2018},
  publisher={Elsevier}
}

@article{behboodi2016mathematical,
  title={A mathematical model for fingerprinting-based localization algorithms},
  author={Behboodi, Arash and Lemic, Filip and Wolisz, Adam},
  journal={arXiv preprint arXiv:1610.07636},
  year={2016}
}

@article{abbasi2016nano,
  title={Nano-communication for biomedical applications: A review on the state-of-the-art from physical layers to novel networking concepts},
  author={Abbasi, Qammer H and Yang, Ke and Chopra, Nishtha and others},
  journal={IEEE Access},
  volume={4},
  pages={3920--3935},
  year={2016},
  publisher={IEEE}
}

@article{stelzner2017function,
  title={Function centric nano-networking: Addressing nano machines in a medical application scenario},
  author={Stelzner, Marc and Dressler, Falko and Fischer, Stefan},
  journal={Nano communication networks},
  volume={14},
  pages={29--39},
  year={2017},
  publisher={Elsevier}
}

@article{abadal2015time,
  title={Time-domain analysis of graphene-based miniaturized antennas for ultra-short-range impulse radio communications},
  author={Abadal, Sergi and Llatser, Ignacio and Mestres, Albert and others},
  journal={IEEE Transactions on Communications},
  volume={63},
  number={4},
  pages={1470--1482},
  year={2015},
  publisher={IEEE}
}

@article{lemic2021survey,
  title={Survey on terahertz nanocommunication and networking: A top-down perspective},
  author={Lemic, Filip and Abadal, Sergi and Tavernier, Wouter and others},
  journal={IEEE Journal on Selected Areas in Communications},
  volume={39},
  number={6},
  pages={1506--1543},
  year={2021},
  publisher={IEEE}
}

@article{dressler2015connecting,
  title={Connecting in-body nano communication with body area networks: Challenges and opportunities of the Internet of Nano Things},
  author={Dressler, Falko and Fischer, Stefan},
  journal={ELSEVIER Nano Communication Networks},
  volume={6},
  number={2},
  pages={29--38},
  year={2015},
  publisher={Elsevier}
}

@article{simonjan2021body,
  title={In-body bionanosensor localization for anomaly detection via inertial positioning and THz backscattering communication},
  author={Simonjan, Jennifer and Unluturk, Bige D and Akyildiz, Ian F},
  journal={IEEE Transactions on NanoBioscience},
  volume={21},
  number={2},
  pages={216--225},
  year={2021},
  publisher={IEEE}
}

@inproceedings{lemic2022toward,
  title={Toward location-aware in-body terahertz nanonetworks with energy harvesting},
  author={Lemic, Filip and Abadal, Sergi and Stevanovic, Aleksandar and Alarc{\'o}n, Eduard and Famaey, Jeroen},
  booktitle={ACM International Conference on Nanoscale Computing and Communication},
  pages={1--6},
  year={2022}
}

@article{lemic2023insights,
  title={Insights from the Design Space Exploration of Flow-Guided Nanoscale Localization},
  author={Lemic, Filip and Bartra, Gerard Calvo and L{\'o}pez, Arnau Brosa and others},
  journal={arXiv preprint arXiv:2305.18493},
  year={2023}
}

@inproceedings{geyer2018bloodvoyagers,
  title={BloodVoyagerS: Simulation of the work environment of medical nanobots},
  author={Geyer, Regine and Stelzner, Marc and others},
  booktitle={ACM International Conference on Nanoscale Computing and Communication},
  pages={1--6},
  year={2018}
}

@article{gomez2023optimizing,
  title={Optimizing terahertz communication between nanosensors in the human cardiovascular system and external gateways},
  author={G{\'o}mez, Jorge Torres and Simonjan, Jennifer and Jornet, Josep Miquel and others},
  journal={IEEE Communications Letters},
  year={2023},
  publisher={IEEE}
}

@article{zhou2020graph,
  title={Graph neural networks: A review of methods and applications},
  author={Zhou, Jie and Cui, Ganqu and Hu, Shengding and others},
  journal={AI open},
  volume={1},
  pages={57--81},
  year={2020},
  publisher={Elsevier}
}

@inproceedings{lymberopoulos2015realistic,
  title={A realistic evaluation and comparison of indoor location technologies: Experiences and lessons learned},
  author={Lymberopoulos, Dimitrios and Liu, Jie and Yang, Xue and others},
  booktitle={IEEE/ACM International Conference on Information Processing in Sensor Networks},
  pages={178--189},
  year={2015}
}

@article{van2015platform,
  title={Platform for benchmarking of RF-based indoor localization solutions},
  author={Van Haute, Tom and De Poorter, Eli and Lemic, Filip and others},
  journal={IEEE Communications Magazine},
  volume={53},
  number={9},
  pages={126--133},
  year={2015},
  publisher={IEEE}
}

@article{wu2020comprehensive,
  title={A comprehensive survey on graph neural networks},
  author={Wu, Zonghan and Pan, Shirui and Chen, Fengwen and others},
  journal={IEEE Transactions on Neural Networks and Learning Systems},
  volume={32},
  number={1},
  pages={4--24},
  year={2020},
  publisher={IEEE}
}

@inproceedings{zhang2019heterogeneous,
  title={Heterogeneous graph neural network},
  author={Zhang, Chuxu and Song, Dongjin and Huang, Chao and others},
  booktitle={ACM SIGKDD international conference on knowledge discovery \& data mining},
  pages={793--803},
  year={2019}
}

@inproceedings{vasisht2018body,
  title={In-body backscatter communication and localization},
  author={Vasisht, Deepak and Zhang, Guo and Abari, Omid and Lu, Hsiao-Ming and Flanz, Jacob and Katabi, Dina},
  booktitle={ACM Special Interest Group on Data Communication},
  pages={132--146},
  year={2018}
}

@inproceedings{pascual2024math,
  title={Mathematical Modelling of Raw Data for Flow-Guided In-body Nanoscale Localization},
  author={Pascual, Guillem and Lemic, Filip and Delgado, Carmen and Costa Perez, Xavier},
  booktitle={IEEE International Conference on Machine Learning for Communication and Networking},
  year={2024}
}

@inproceedings{simeoni2020long,
  title={Long-rangeultrasound wake-up receiver with a piezoelectric nanoscale ultrasound transducer (pNUT)},
  author={Simeoni, Pietro and Castellani, Matteo and Piazza, Gianluca},
  booktitle={International Conference on Micro Electro Mechanical Systems},
  pages={849--852},
  year={2020},
  organization={IEEE}
}

@article{chen2018deep,
  title={Deep-subwavelength control of acoustic waves in an ultra-compact metasurface lens},
  author={Chen, Jian and Xiao, Jing and Lisevych, Danylo and Shakouri, Amir and Fan, Zheng},
  journal={Nature communications},
  volume={9},
  number={1},
  pages={4920},
  year={2018},
  publisher={Nature Publishing Group UK London}
}

@article{vizziello2023intra,
  title={Intra-body communications for nervous system applications: Current technologies and future directions},
  author={Vizziello, Anna and Magarini, Maurizio and Savazzi, Pietro and others},
  journal={Computer Networks},
  volume={227},
  pages={109718},
  year={2023},
  publisher={Elsevier}
}

@inproceedings{engstrand2022end,
  title={End-to-end transmission of physiological data from implanted devices to a cloud-enabled aggregator using fat intra-body communication in a live porcine model},
  author={Engstrand, Johan and Perez, Mauricio and others},
  booktitle={European Conference on Antennas and Propagation},
  pages={1--5},
  year={2022},
  organization={IEEE}
}

@article{khaleghi2018radio,
  title={Radio frequency backscatter communication for high data rate deep implants},
  author={Khaleghi, Ali and Hasanvand, Aminolah and Balasingham, Ilangko},
  journal={IEEE Transactions on Microwave Theory and Techniques},
  volume={67},
  number={3},
  pages={1093--1106},
  year={2018},
  publisher={IEEE}
}

@article{geurts2016temporal,
  title={Temporal profile of body temperature in acute ischemic stroke: relation to infarct size and outcome},
  author={Geurts, Marjolein and Scheijmans, F{\'e}line EV and van Seeters, Tom and Biessels, Geert J and Kappelle, L Jaap and Velthuis, Birgitta K and van der Worp, H Bart and DUST investigators},
  journal={BMC neurology},
  volume={16},
  pages={1--7},
  year={2016},
  publisher={Springer}
}

@article{zhang2015dna,
  title={DNA methylation consistency implicates the primary tumor cell origin of recurrent hepatocellular carcinoma},
  author={Zhang, Xiaolei and Liu, Shuang and Shen, Congle and Wu, Yali and Zhang, Ling and Chen, Xiangmei and Lu, Fengmin},
  journal={Epigenomics},
  volume={7},
  number={4},
  pages={581--592},
  year={2015},
  publisher={Future Medicine}
}

@inproceedings{wendt2023proteome,
  title={Proteome Fingerprinting as a Localization Scheme for Nanobots},
  author={Wendt, Regine and Lau, Florian-Lennert and Unger, Lena and Fischer, Stefan},
  booktitle={Proceedings of the 10th ACM International Conference on Nanoscale Computing and Communication},
  pages={27--32},
  year={2023}
}

@inproceedings{hu2020heterogeneous,
  title={Heterogeneous graph transformer},
  author={Hu, Ziniu and Dong, Yuxiao and Wang, Kuansan and Sun, Yizhou},
  booktitle={Proceedings of the web conference},
  pages={2704--2710},
  year={2020}
}

@inproceedings{von2023transformers,
  title={Transformers learn in-context by gradient descent},
  author={Von Oswald, Johannes and Niklasson, Eyvind and Randazzo, Ettore and others},
  booktitle={International Conference on Machine Learning},
  pages={35151--35174},
  year={2023},
  organization={PMLR}
}

@inproceedings{jiang2019gaussian,
  title={Gaussian-induced convolution for graphs},
  author={Jiang, Jiatao and Cui, Zhen and Xu, Chunyan and Yang, Jian},
  booktitle={Proceedings of the AAAI conference on artificial intelligence},
  volume={33},
  number={01},
  pages={4007--4014},
  year={2019}
}

\end{document}